\documentclass[final,5p,times,twocolumn]{elsarticle}

\usepackage{graphicx}
\usepackage{amssymb}
\usepackage{latexsym}
\usepackage{lipsum}
\usepackage{amsmath}
\usepackage{multicol}
\usepackage{diagbox}
\usepackage{mathtools}
\usepackage{hyperref}
\usepackage{makecell}
\usepackage{subfigure}
\usepackage{framed,multirow}
\usepackage[ruled,vlined]{algorithm2e}
\usepackage{float}
\usepackage{url}
\usepackage{multirow}
\hypersetup{colorlinks = true}
\usepackage{amsmath}
\DeclareMathOperator*{\argmin}{argmin}

\SetCommentSty{mycommfont}

\usepackage[table]{xcolor}
\definecolor{newcolor}{rgb}{.8,.349,.1}

\usepackage{natbib}
\biboptions{semicolon,round}

\setcitestyle{authoryear,open={(},close={)}}

\hypersetup{
     colorlinks   = true,
     citecolor    = blue
}
\usepackage{etoolbox}
\makeatletter
\patchcmd{\ps@pprintTitle}
  {Preprint submitted to}
  {To appear in}
  {}{}
\makeatother

\journal{NeuroImage}

\begin{document}

\begin{frontmatter}


\title{Longitudinal diffusion MRI analysis using Segis-Net: a single-step deep-learning framework for simultaneous segmentation and registration}

\author[2]{Bo Li\corref{cor1}}
\ead{b.li@erasmusmc.nl}
\cortext[cor1]{Corresponding author}
\author[2,3]{Wiro J. Niessen}
\author[2]{Stefan Klein}
\author[2,4]{Marius de Groot}
\author[2,4,5]{M. Arfan Ikram}
\author[2,4]{Meike W. Vernooij}
\author[2]{Esther E. Bron}

\address[2]{Department of Radiology and Nuclear Medicine, Erasmus MC, Rotterdam, the Netherlands}
\address[3]{Imaging Physics, Applied Sciences, Delft University of Technology, the Netherlands}
\address[4]{Department of Epidemiology, Erasmus MC, Rotterdam, the Netherlands}
\address[5]{Department of Neurology, Erasmus MC, Rotterdam, the Netherlands}

\date{}

\begin{abstract}
This work presents a single-step deep-learning framework for longitudinal image analysis, coined Segis-Net. To optimally exploit information available in longitudinal data, this method concurrently learns a multi-class segmentation and nonlinear registration. Segmentation and registration are modeled using a convolutional neural network and optimized simultaneously for their mutual benefit. An objective function that optimizes spatial correspondence for the segmented structures across time-points is proposed. We applied Segis-Net to the analysis of white matter tracts from N=8045 longitudinal brain MRI datasets of 3249 elderly individuals. Segis-Net approach showed a significant increase in registration accuracy, spatio-temporal segmentation consistency, and reproducibility compared with two multistage pipelines. This also led to a significant reduction in the sample-size that would be required to achieve the same statistical power in analyzing tract-specific measures. Thus, we expect that Segis-Net can serve as a new reliable tool to support longitudinal imaging studies to investigate macro- and microstructural brain changes over time.
\end{abstract}



\end{frontmatter}



\newcommand\blfootnote[1]{%
  \begingroup
  \renewcommand\thefootnote{}\footnote{#1}%
  \addtocounter{footnote}{-1}%
  \endgroup
}
\blfootnote{\textit{Abbreviations}: MRI, Magnetic Resonance Imaging; DTI, Diffusion Tensor Imaging; FA, Fractional Anisotropy; MD, Mean Diffusivity; TE, Echo Time; TR, Repetition Time}

\newif\ifcitation
\citationtrue
\DeclareRobustCommand{\Vv}[1]{\ifcitation Van\else van\fi}
\DeclareRobustCommand{\Dd}[1]{\ifcitation De\else de\fi}

\newcommand{\beginsupplement}{%
        \setcounter{table}{0}
        \renewcommand{\thetable}{S\arabic{table}}%
        \setcounter{figure}{0}
        \renewcommand{\thefigure}{S\arabic{figure}}%
     }

\section{Introduction}
\label{S:1}
The increasing availability of longitudinal imaging data is expanding our ability to capture and characterize progressive anatomical changes, ranging from normal changes in the life span, to responses along disease trajectories or therapeutic actions. Compared to cross-sectional studies, longitudinal imaging studies have the advantage of allowing to trace the order of events at the individual level and to correct for the confounding effect of time-invariant individual differences \citep{van2017temporal}. They are thus considered to be more accurate and sensitive in capturing subtle changes over time. To analyze spatio-temporal changes from longitudinal imaging data, a tailored framework that involves both segmentation and registration is required to segment the structures-of-interest and to register temporal frames. This can be achieved by directly combining two existing segmentation and registration tools, which are often designed for cross-sectional studies. However, the information offered in longitudinal data remains underutilized.

Various studies have shown that combining segmentation and registration at the stage of algorithm optimization can lead to improved performance. A popular combination strategy is to use the output of one task to optimize the other. Registration can be improved by using segmentation-level correspondences as input for deformation initialization \citep{dai1999feature,postelnicu2008combined} and optimization \citep{de2013improving,rohe2017svf,hu2018label,balakrishnan2019voxelmorph,bastiaansen2020towards,zhu2020neurreg}. Likewise, segmentation can benefit from registration by propagating anatomical information to subsequent frames, as has been shown in classical multi-atlas based segmentation methods \citep{fischl2002whole,vakalopoulou2018atlasnet} and in recent data-augmentation techniques which introduce labels to support unsupervised \citep{pathak2017learning} and weakly-supervised segmentation \citep{bortsova2019semi,vlontzos2018deep}.

Other approaches combine the optimization of parameters from both tasks on a deeper level. \cite{wyatt2003map} sub-grouped these methods into two types according to the way in which they update their parameters: 1) ``simultaneous estimation" that updates both the class labels and the transformations in a single-step optimization, and 2) ``joint estimation" that alternately updates (separate) models in a multi-step optimization. Although the initialization and robustness of joint estimation can be influenced by the selection of the order to optimize and the criteria to switch tasks, this approach is preferred as it requires less computation power and allows to use task-specific training datasets \citep{yezzi2003variational,wyatt2003map,ashburner2005unified,pohl2006bayesian,parisot2014concurrent,gooya2011joint,cheng2017segflow,xu2019deepatlas}. Simultaneous estimation is expected to be more accurate, as it fully exploits the conditional correlations between two tasks that can be discounted in sequential processing \citep{ashburner2005unified}. In addition, simultaneous estimation can explicitly optimize performances that rely on both tasks. We expect that this advantage has a large potential in improving the reliability of analysis of longitudinal imaging data, for instance by optimizing the spatio-temporal consistency of the segmentation. With the growing capability of modeling and computation by deep learning techniques, several simultaneous methods have been proposed and coupled segmentation with deformable registration in different ways, either for 2D \citep{qin2018joint} or 3D images \citep{li2019hybrid,estienne2019u,estienne2020deep}.

Diffusion magnetic resonance imaging (MRI) is a non-invasive imaging technique that measures the diffusion of water in-vivo and can be used to quantitatively characterize white matter (WM) microstructure. In addition, diffusion MRI derived measures, such as diffusion tensor imaging (DTI) metrics \citep{le2001diffusion}, are likely to be more sensitive than structural measures in the early detection of changes in WM, and are therefore promising for the identification of subtle changes that relate to the early stages of the disease \citep{niessen2016mr}, for instance in studying dementia subtypes \citep{meijboom2019exploring}. Longitudinal diffusion MRI has been widely studied at various levels, i.e., from regions-of-interest \citep{sullivan2010longitudinal,keihaninejad2013unbiased}, to tract level \citep{lebel2011longitudinal, yendiki2016joint,meijboom2019exploring,dimond2020early}, and voxel level \citep{barrick2010white,farbota2012longitudinal,de2016white}. Since WM tracts are functionally grouped axonal fibers and thought to subserve particular brain functions, tract-specific investigation may highlight categorical differences in vulnerability to neurodegeneration and bridge the interpretation of imaging biomarkers with clinical symptoms. 

Segmentation of WM tracts is however non-trivial because tracts cannot be identified directly from diffusion MRI, i.e., there is no in-vivo ``gold standard'' for tract \citep{crick1993backwardness}, and because their anatomy can be complex. WM tracts are commonly segmented based on diffusion tractography by reconstruction of potential fiber pathways \citep{conturo1999tracking}. Recently, deep learning based methods, in particular using convolutional neural networks (CNN), have emerged and showed promising accuracy and efficiency in segmenting WM tracts \citep{li2018reproducible,li2020neuro4neuro,wasserthal2018tractseg}. 

In the present work we focus on a CNN-based framework for longitudinal analysis of WM tracts, i.e., Segis-Net, and investigate the value of simultaneous optimization of segmentation and registration in this setting. In \cite{li2019hybrid}, we introduced a generic framework for simultaneous optimization, in which increased accuracies of both tasks were observed in a pilot analysis of a single tract (forceps minor; FMI). In this paper, we extend the tract-specific method by enabling concurrent segmentation of multiple tracts, which is a non-trivial task as a voxel can belong to multiple tracts. This also solves the problem of inconsistencies in deformations because of tract-specific ROIs. The registration task within the framework is updated to learn only local deformations rather than an end-to-end composite including rigid transformation, as brain local changes over time is a focus in longitudinal imaging studies. In addition, we compare the performance of Segis-Net to two multistage pipelines based on both classical and deep learning algorithms, and two state-of-the-art methods. The segmentation accuracy, registration accuracy, spatio-temporal consistency of segmentation, and reproducibility of segmentation and tract-specific measures of the pipelines are quantitatively evaluated. Also, we evaluate the sample-size reduction that can be achieved in the imaging analysis of WM tracts to provide insight into the practical value of the methods in clinical applications. 

\section{Methods}
\label{S:2}
In this section, we first describe how the segmentation and registration tasks are individually modeled using CNN-based approaches. Subsequently, we present the proposed Segis-Net that integrates both tasks in a single-step CNN framework.

\subsection{CNN-based image segmentation}
\label{ss:2.1}
Given a $n$-D image $I$ which can be described by either intensity values, multi-channel features or directional tensors, the goal of CNN-based segmentation is to automatically infer, for each voxel $x \in \mathbb{R}^n$, its probability of belonging to the structure $k \in [1,K]$ with $K$ the number of structures, i.e., voxel-wise classification. The CNN model can be interpreted as a parameterized mapping function $\mathcal{F}_{\boldsymbol{\Theta}}$ such that the segmented structures spatially correspond to a segmentation ground truth with multiple channels $\mathcal{S}=\{S_1,...,S_K\}$. The estimation of the segmentation is formulated as:
\begin{equation}\label{eq1}
 \hat{\mathcal{S}} = \mathcal{F}_{\boldsymbol{\Theta}}\big(I\big).
\end{equation} 
$\mathcal{F}_{\boldsymbol{\Theta}}$ is commonly modeled by a nested series of convolutions, non-linearity, normalization, and re-sampling operations embedded in the network architecture. $\boldsymbol{\Theta}$ indicate trainable parameters. 

The procedure of estimating parameter $\boldsymbol{\Theta}$ is then defined as an optimization with respect to a loss function $\mathcal{L}_{seg}$, aiming at minimizing the classification error over all the $N$ pairs of training samples $\{(\mathcal{S}^i, I^i)\}_{i=1}^N$, i.e.,
\begin{equation}\label{eq2}
    \boldsymbol{\Theta} \leftarrow \argmin_{\boldsymbol{\Theta}} \sum_{i=1}^N \mathcal{L}_{seg}\, \bigg(\mathcal{S}^i, \mathcal{F}_{\boldsymbol{\Theta}}(I^i)\bigg).
\end{equation}

The loss function comprises metrics that quantify the difference between the prediction and the ground truth. In this study, $\mathcal{L}_{seg}$ is the average Dice coefficient \citep{dice1945measures,crum2006generalized} over all $K$ structures:
\begin{equation}\label{eq3}
    \mathcal{L}_{seg} (\mathcal{S}, \hat{\mathcal{S}} ) = - \frac{2}{K} \sum_{k=1}^K \frac{\sum_x S_k \hat{S}_k}{\sum_x (S_k)^2 + \sum_x (\hat{S_k})^2}.
\end{equation}

After estimation of the map function $\mathcal{F}_{\boldsymbol{\Theta}}$, a probabilistic prediction for the structures of interest $\hat{\mathcal{S}}$ in a given image can be inferred (Eq.\ref{eq1}).

\subsection{CNN-based deformable registration}
\label{ss:2.2}

Let us consider a pair of $n$-D images, $I_s$ in the source space $\Omega\,_s \subset \mathbb{R}^n$, and $I_t$ in the target space $\Omega\,_t \subset \mathbb{R}^n$, which contain a common structure to be aligned. The spatial correspondence between images can be established by estimating a dense displacement field $\boldsymbol{\phi}$, such that $I_s \circ \boldsymbol{\phi}$ and $I_t$ correspond spatially.

In line with the hierarchical optimization scheme of classical registration algorithms, most existing learning-based registration methods use affine alignment as a prepossessing step, in which case the displacement field denotes the composition of affine and deformable transform, i.e., $\boldsymbol{\phi} = \boldsymbol{\phi}_A \circ \boldsymbol{\phi}_D$. To estimate $\boldsymbol{\phi}_D$, the CNN model can be interpreted as a shared domain-invariant mapping function $\mathcal{G}_{\boldsymbol{\Psi}}$ such that for any unseen pair of images a most likely transformation between them can be inferred without pair-specific optimization, i.e., 
\begin{equation}\label{eq4}
\hat{\boldsymbol{\phi}}_D = \mathcal{G}_{\boldsymbol{\Psi}}(I_t,\, I_s \circ \boldsymbol{\phi}_A) 
\end{equation}
\begin{equation}\label{eq5}
\implies \quad I_s \circ \hat{\boldsymbol{\phi}} \leftarrow \mathcal{G}_{\boldsymbol{\Psi}}(I_t,\, I_s,\, \boldsymbol{\phi}_A).
\end{equation}

The parameters $\boldsymbol{\Psi}$ of the mapping function are optimized based on a registration dissimilarity loss $\mathcal{L}_{reg}$, aimed at minimizing the registration error. Meanwhile, to penalize large deviations of deformation and preserve anatomical topology during transformations, a deformation smoothness term $\mathcal{L}_{def}$ is commonly included in the loss function. In this work, we use the mean squared error based on intensities for $\mathcal{L}_{reg}$ and the average spatial gradients of the displacement field for $\mathcal{L}_{def}$, i.e., 
\begin{equation}\label{eq6}
     \mathcal{L}_{reg}\, (I_t,\, I_s \circ \hat{\boldsymbol{\phi}}) =  \frac{1}{|\Omega\,_t|}\, \, \|I_t - I_s \circ \hat{\boldsymbol{\phi}} \| \,_2^2,
\end{equation}
\begin{equation}\label{eq7}
     \mathcal{L}_{def}\, (\hat{\boldsymbol{\phi}}_D) = \frac{1}{|\Omega\,_t|}\,  \|\nabla \hat{\boldsymbol{\phi}}_D \|\, _2^2.
\end{equation}
Combining Eq.~(\ref{eq4}), (\ref{eq5}), (\ref{eq6}) and (\ref{eq7}), the estimation of $ \boldsymbol{\Psi}$ over all the $N$ training samples $\{(I_t^i, I_s^i, \boldsymbol{\phi}_A^i)\}_{i=1}^N$ can be formulated as: 
\begin{equation}\label{eq8}
    \boldsymbol{\Psi} \leftarrow \argmin_{\boldsymbol{\Psi}} \sum_{i=1}^N \mathcal{L}_{reg} \big(I_t^i,\, \mathcal{G}_{\boldsymbol{\Psi}}(I_t^i,\, I_s^i,\, \boldsymbol{\phi}_A^i)\big) + \mathcal{L}_{def} \big(\mathcal{G}_{\boldsymbol{\Psi}}(I_t^i,\, I_s^i \circ \boldsymbol{\phi}_A^i)\big).
\end{equation}

\subsection{Simultaneous estimation of segmentation and registration}
\label{ss:2.3}

\begin{figure*}[!t]
\centering
\includegraphics[width=\textwidth]{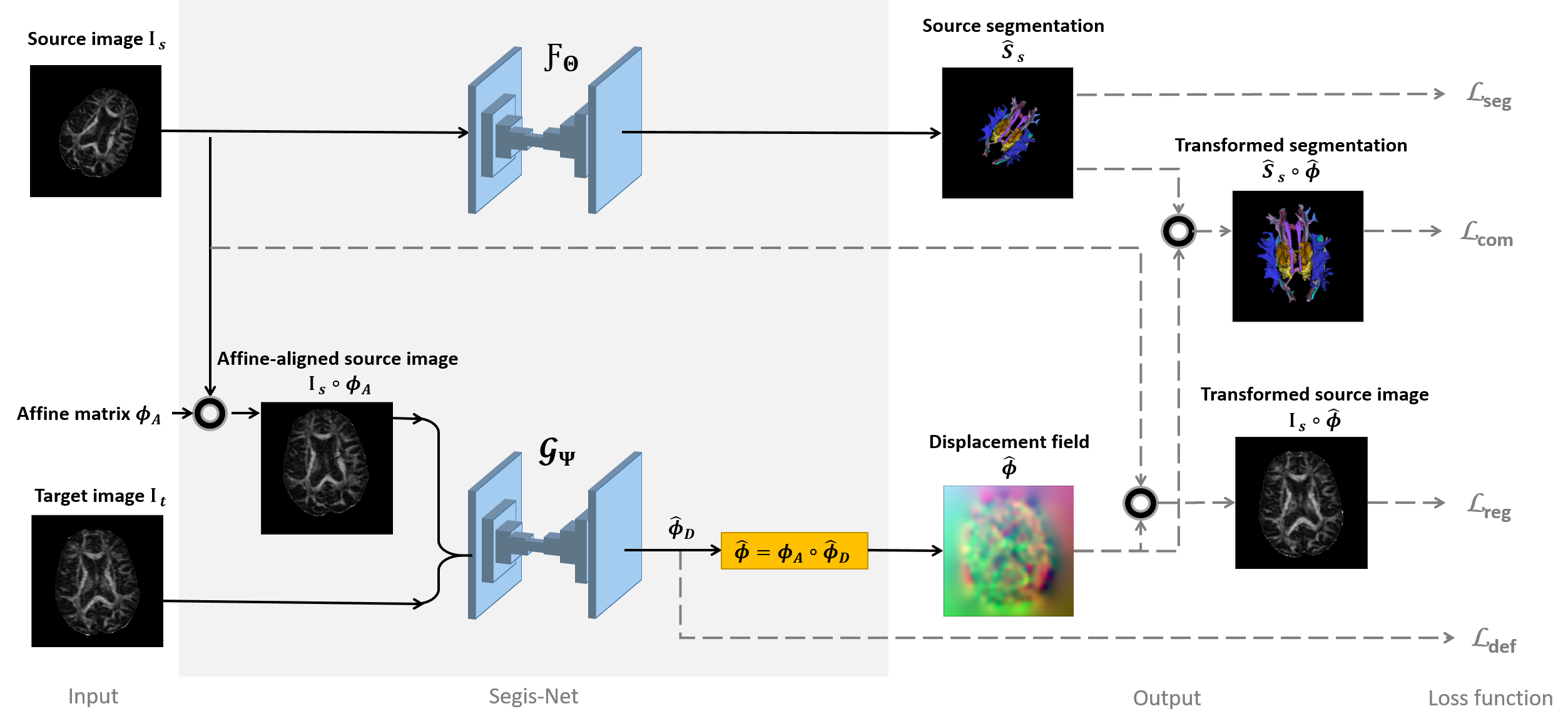}
\caption{Overview of the Segis-Net framework. $\boldsymbol{\Theta}$ and $\boldsymbol{\Psi}$ denote the parameters of the segmentation ($\mathcal{F}_{\boldsymbol{\Theta}}$) and registration ($\mathcal{G}_{\boldsymbol{\Psi}}$) function, respectively. Black circle indicates spatial warp with affine matrix ($\boldsymbol{\phi}_A$) or the composite displacement field ($\hat{\boldsymbol{\phi}}$). The concatenation of the affine-aligned images is used as the input for $\mathcal{G}_{\boldsymbol{\Psi}}$. Loss function consists of $\mathcal{L}_{seg}$, $\mathcal{L}_{com}$, $\mathcal{L}_{reg}$ and $\mathcal{L}_{def}$ terms. Solid lines indicate the primary workflow of the method; dashed lines indicate the operations that are only implemented during training and could be adapted for applications.}\label{fig1}
\end{figure*}

In this work, we aim to simultaneously estimate the parameters for segmentation ($\boldsymbol{\Theta}$) and for registration ($\boldsymbol{\Psi}$) in a single-step optimization. For this purpose, we integrate the segmentation and registration function $\mathcal{F}_{\boldsymbol{\Theta}}$ and $\mathcal{G}_{\boldsymbol{\Psi}}$ using an end-to-end optimization with the Segis-Net. The loss function of the Segis-Net is designed to meet the joint objective of both tasks and meanwhile to optimize the spatio-temporal consistency of segmentation which rely on both tasks. The overview of the proposed framework is illustrated in Fig.~\ref{fig1}. We describe the framework architecture and loss function in the following paragraphs.

\subsubsection{Segis-Net framework}
\label{sss:2.3.1}

In the present study, we focus on the analysis of 3D images and utilize 3D convolutions for the Segis-Net framework. The framework involves function $\mathcal{F}_{\boldsymbol{\Theta}}$ and $\mathcal{G}_{\boldsymbol{\Phi}}$ as two parallel streams that interact on their outputs. In order to eliminate the loss in image quality caused by multiple interpolations, Segis-Net warps source images with only the composite displacement fields ($\boldsymbol{\phi}$) by taking as input the original source image ($I_s$) and pre-estimated affine matrix ($\boldsymbol{\phi}_A$). This design has additional advantages over existing methods that prepare all ordered pairs of affine-aligned images in disk storage, as only up to half the storage is needed and as it can be flexibly applied to related images in the same space such as the DTI metrics.  

$\mathcal{F}_{\boldsymbol{\Theta}}$ outputs a set of probabilistic segmentations ($\hat{\mathcal{S}}_s$) of the source image. $\mathcal{G}_{\boldsymbol{\Psi}}$ outputs a dense local displacement $\hat{\boldsymbol{\phi}}_D$ along the x, y, and z axes. The source image and its segmentations are subsequently warped into the target space using the composed displacement field. The warp operation is implemented by a computational layer with differentiable trilinear interpolation \citep{ jaderberg2015spatial,balakrishnan2019voxelmorph}. The segmentation and registration streams have independent network architectures which are only connected by the output, i.e., the transformed source-segmentation to the target space ($\hat{\mathcal{S}}_s \circ \hat{\boldsymbol{\phi}}$). Thus, they can be applied separately after taking advantage of the simultaneous optimization. The Segis-Net framework gives four outputs during training: 
\begin{enumerate}
\item The segmentation of the structures of interest from the source image ($\hat{\mathcal{S}}_s$),
\item A local displacement field between the source and target images ($\hat{\boldsymbol{\phi}}_D$),
\item The warped source image in the target space ($I_s \circ \hat{\boldsymbol{\phi}}$),
\item The warped source segmentations in the target space ($\hat{\mathcal{S}}_s \circ \hat{\boldsymbol{\phi}}$).
\end{enumerate}

We propose a generic framework where the architecture of each stream can be adapted based on specific applications. For the particular network used in this study, we encoded two streams with a U-Net architecture, that was modified as detailed below \citep{ronneberger2015u}. In short, each stream was composed of an encoder and decoder path with skip connections of feature pyramid at multiple scales in order to merge coarse- and fine-convolved features, similar to the multi-resolution strategy used in classical algorithms to increase robustness. The encoder paths with max-pooling operation between convolution layers gradually extract abstract features for the target anatomy ($\mathcal{F}_{\boldsymbol{\Theta}}$) and global transformation between images ($\mathcal{G}_{\boldsymbol{\Psi}}$). Subsequently, the decoder paths restore the details in segmentations ($\mathcal{F}_{\boldsymbol{\Theta}}$) and refine local deformations ($\mathcal{G}_{\boldsymbol{\Psi}}$) by linear up-sampling the feature maps and concatenating them with the coarse counterpart at the same scale. The convolution layers produce a set of feature maps by individually convolving inputs with 3D kernels of size $(3,3,3)$, followed by batch normalization \citep{ioffe2015batch} and a leaky ReLu layer ($a=0.2$) for modeling non-linearity \citep{maas2013rectifier}. For the segmentation stream ($\mathcal{F}_{\boldsymbol{\Theta}}$), we split the output layer into sub-branches to facilitate multi-class classification for voxels with multiple labels. The final layer of the sub-branches consisted of a $(1,1,1)$ convolution and a sigmoid activation. For the registration stream, the output layer was a convolution layer with three kernels that yielded the local displacement $\hat{\boldsymbol{\phi}}_D$. We provide detailed implementation of the network architecture in the supplementary material (Figure \ref{fig:s1} and \ref{fig:s2}).

\subsubsection{Segis-Net loss function}
\label{sss:2.3.2}

The loss function of Segis-Net is composed of four terms that measure segmentation accuracy ($\mathcal{L}_{seg}$, Eq.~\ref{eq3}), intensity similarity between registered images ($\mathcal{L}_{reg}$, Eq.~\ref{eq6}), deformation field smoothness ($\mathcal{L}_{def}$, Eq.~\ref{eq7}), and longitudinal composite of registration and segmentation ($\mathcal{L}_{com}$, Eq.~\ref{eq11}). It is formulated as:
\begin{equation}\label{eq9}
    \begin{split} 
      \mathcal{L} & = \mathcal{L}_{seg} \big(\mathcal{S}_s,\, \mathcal{\hat{S}}_s\big)\\
     & \quad +\alpha \mathcal{L}_{reg} \big(I_t,\, I_s \circ \hat{\boldsymbol{\phi}})\big)
     +\, \beta \mathcal{L}_{def} \big(\hat{\boldsymbol{\phi}}_D\big) \\
    & \quad  +\gamma \mathcal{L}_{com} \big(\mathcal{S}_t,\, \hat{\mathcal{S}}_s \circ \hat{\boldsymbol{\phi}}\big),
    \end{split} 
\end{equation}
and optimized for $\boldsymbol{\Theta}$ and $\boldsymbol{\Psi}$ over all $N$ training samples $\{(\mathcal{S}^i_t, \mathcal{S}^i_s, I_t^i, I_s^i, \boldsymbol{\phi}_A^i)\}_{i=1}^N$:
\begin{equation}\label{eq10}
    \begin{split} 
        \boldsymbol{\Theta}, \boldsymbol{\Psi} & \leftarrow \argmin_{ \boldsymbol{\Theta},\boldsymbol{\Psi}} \sum_{i=1}^N \mathcal{L}_{seg}\, \big(\mathcal{S}^i_s, \mathcal{F}_{\boldsymbol{\Theta}}(I^i_s)\big)\\
      & \quad + \alpha \mathcal{L}_{reg} \big(I_t^i,\, \mathcal{G}_{\boldsymbol{\Psi}}(I_t^i,\, I_s^i,\, \boldsymbol{\phi}_A^i)\big) + \beta \mathcal{L}_{def} \big(\mathcal{G}_{\boldsymbol{\Psi}}(I_t^i,\, I_s^i \circ \boldsymbol{\phi}_A^i)\big)\\
      & \quad + \gamma \mathcal{L}_{com}\big( \mathcal{S}^i_t, \mathcal{F}_{\boldsymbol{\Theta}}(I_s^i), \mathcal{G}_{\boldsymbol{\Psi}}(I_t^i,\, I_s^i,\, \boldsymbol{\phi}_A^i)\big).
    \end{split}
\end{equation}

We quantify the longitudinal composite loss term using the average Dice coefficient over all $K$ structures:

\begin{equation}\label{eq11}
    \mathcal{L}_{com} \big(\mathcal{S}_t,\, \hat{\mathcal{S}}_s \circ \hat{\boldsymbol{\phi}}\big) = - \frac{2}{K} \sum_{k=1}^K \frac{ \sum_{x \in \Omega_t} S_t^k  (\hat{S}_s^k \circ \hat{\boldsymbol{\phi}} )}{\sum_{x \in \Omega_t}(S_t^k)^2 + \sum_{x \in \Omega_t} (\hat{S}_s^k \circ \hat{\boldsymbol{\phi}})^2}.
\end{equation}
In longitudinal imaging studies, the spatial correspondence in segmentation depends on the performance of both the segmentation and the registration procedure. Besides an explicit optimization of correspondence, the $\mathcal{L}_{com}$ term also exploits longitudinal information to boost both tasks, which introduces some degree of augmentation and regularization for registration and on the other hand constraints and prior knowledge for segmentation. 

The hyperparameters $\alpha$ and $\gamma$ balance the loss magnitude of segmentation, registration, and their interdependent composite. The degree of regularization on the deformation is described by $\beta$. The procedure of simultaneous optimization is summarized with pseudo code in supplementary material (Algorithm \ref{algorithm1}).

\section{Application to diffusion MRI}
\label{s:3}

The performance of Segis-Net is demonstrated by analyzing white matter tracts in a large diffusion MRI dataset, and compared to that of two multi-stage pipelines, in which segmentation and registration are independently optimized. Performance is evaluated in a longitudinal setting where multiple time-points from the same individual are available.
 
\subsection{Dataset} \label{ss:3.1}
\begin{figure}[!ht]
\centering
\includegraphics[scale=0.6]{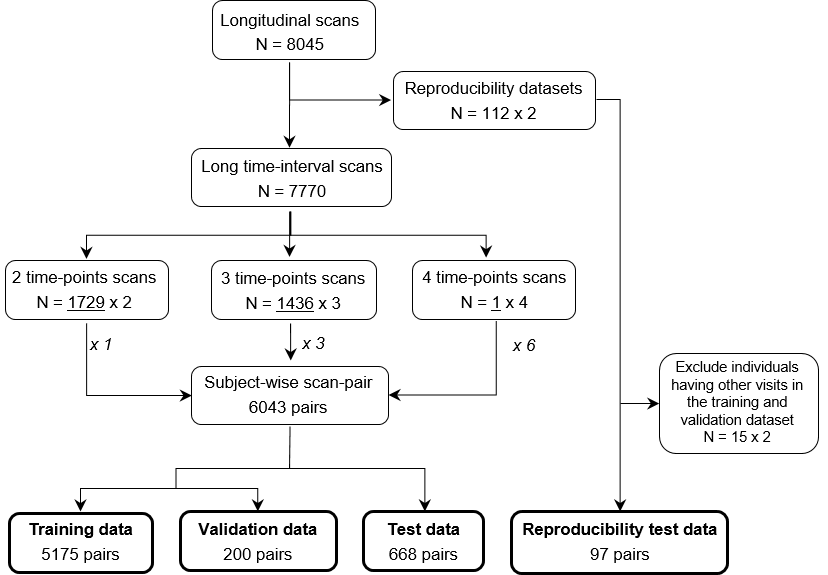}
\caption{A flowchart for the inclusion, exclusion, and split of the datasets.}\label{figx}
\end{figure}

The Rotterdam Study is a prospective and population-based study targeting causes and consequences of age-related diseases \citep{ikram2020objectives}. For the present analysis, we included 3249 individuals who underwent diffusion MRI scanning twice or more often, resulting in $N=8045$ scans. The mean age at first scan was $61.2 \pm 9.4$ years (range: $45.7 - 91.1$ years). The number of female participants was 1780 ($54.8\%$). A flowchart for the inclusion, exclusion, and split of the datasets is shown in Figure~\ref{figx}. We split the data into two subsets. The larger subset was repeatedly acquired in a time interval of 1--5 years ($N=7770$ scans from 3166 individuals). In these long time-interval scans, it is expected that brain microstructure changes due to aging exist. By matching any two time-points from the same individual regardless of the visiting order, these long time-interval scans can be grouped into 6043 pairs. We used 5175 pairs of scans as training data, 200 pairs as validation data to tune the hyperparameters, monitor the decay of learning rate and select the optimal epoch, and used an independent cohort of 668 pairs for testing. The remaining scans from the smaller subset were from 97 individuals who were scanned twice within a month. No changes in brain macro- and microstructure were expected within such a short time-interval. We used these scans for evaluation of reproducibility of the algorithm. The data split was based on the participants, namely, we made sure that scans from the same participant ended up in either training, validation, or test dataset.

\subsection{MRI acquisition} \label{ss:3.2}
Scans were acquired on a 1.5T MRI scanner (GE Signa Excite). The acquisition parameters for structural and diffusion MRI can be found in \cite{ikram2011rotterdam}. Specifically, diffusion MRI was scanned with the following parameters: TR/TE =$ 8575 ms/82.6 ms$, imaging matrix of $64\times96$, FOV=$21\times21 cm^2$, 35 contiguous slices with slice thickness 3.5 mm, 25 diffusion weighted volumes with a b-value of $1000 s/mm^2$ and 3 non-weighted volumes (b-value $=0 s/mm^2$). The voxel size was resampled from $3.3\times2.2\times3.5 mm^3$ to $1 mm^3$ as required for probabilistic tractography \citep{behrens2007probabilistic}.

\subsection{Image preprocessing} \label{ss:3.3}

Diffusion data were preprocessed using a standardized pipeline \citep{koppelmans2014global}. In short, motion and eddy currents were corrected by affine co-registration of all diffusion weighted volumes to the averaged b0 volumes, including correction of gradient vector directions using Elastix software \citep{klein2010elastix}. Diffusion tensors were estimated with a Levenberg–Marquardt non-linear least-squares optimization algorithm  \citep{leemans2009exploredti}. We subsequently computed DTI measures: fractional anisotropy (FA) and mean diffusivity (MD). Due to noise, tensor estimation failed in a small proportion of voxels, resulting in significant outliers. Outlier voxels with a tensor norm (Frobenius norm) larger than $0.1 mm^2/s$ were set to zero \citep{zhang2007high}. Brain tissue masks including WM and gray matter segmentations were obtained based on structural imaging \citep{vrooman2007multi} and applied to the diffusion tensor images. In this study, we used a ROI of $112 \times 208 \times 112$ voxels to analyze six WM tracts, including left and right cingulate gyrus part of cingulum (CGC), left and right parahippocampal part of cingulum (CGH), forceps major (FMA) and forceps minor (FMI). Diffusion tensor images were image-wise normalized by setting the union of the six components to zero-mean and standard deviation of one. The affine matrix ($\boldsymbol{\phi}_A$) of each image pair was estimated by optimizing the mutual information of FA images using Elastix software.

\subsection{Reference segmentations} \label{ss:3.4}
 
The segmentation labels for model training and evaluation were generated using a probabilistic tractography and atlas-based segmentation method by \cite{de2015tract}. The resulting tract-density images for each tract were normalized by division with the total number of tracts in the tractography run. Finally, tract-specific thresholds for the normalized density images were established by maximizing the reproducibility of FA measures on a subset of 30 participants \citep{de2013improving}. We did not exclude this subset from the reproducibility test data (Figure~\ref{figx}), as it remains unseen to the proposed method and other baseline methods.

\subsection{Baseline multi-stage pipelines} \label{ss:3.5}
We compared the performance of the proposed Segis-Net with two multi-stage pipelines that consist of either non-learning-based or learning-based algorithms to investigate the added value of simultaneous optimization. To assess whether the performance difference between approaches was statistically significant, paired t-tests with P-value threshold $< 0.05$ and Bonferroni correction for controlling the family-wise error of multiple testings were performed. 

First, a non-learning-based \textit{Classical} pipeline was built using an existing tractography-based segmentation algorithm (Section \ref{ss:3.4}) and a deformable registration algorithm Elastix \citep{klein2010elastix}. Elastix was adopted as a competing classical registration method since it has been widely used on our dataset and thereby an optimal parameter setting can be applied for performance comparison. Elastix is designed to run in a cascade of resolutions, and offers the choice between multiple objective functions and multiple optimizers including an efficient adaptive stochastic gradient descent optimizer \citep{klein2009adaptive}. For Elastix (version 4.8), we used a rigid, affine, and B-spline transformation model consecutively by maximizing mutual information between images. The B-spline transformation of spline order 3 was implemented using a multi-resolution framework with isotropic control-point spacing of 24, 12, and 6 $mm$ in three-level resolutions. The maximum number of iterations was 1024. 

Second, we built a learning-based \textit{CNN} pipeline using components from the proposed Segis-Net to evaluate the sole contribution of simultaneous optimization. In this pipeline, we split the integrated segmentation $\mathcal{F}_{\boldsymbol{\Theta}}$ and registration stream $\mathcal{G}_{\boldsymbol{\Phi}}$ into two separate neural networks for independent optimization. Subsequently, the segmented images and estimated transformations were combined. The segmentation network had the same architecture as that for $\mathcal{F}_{\boldsymbol{\Theta}}$, except being independently optimized using the segmentation accuracy $\mathcal{L}_{seg}$ term. As this is a typical setting for CNN-based segmentation approaches \citep{li2018reproducible,ronneberger2015u}, we denote it as \textit{Seg-Net}. Similarly, the registration network denoted as \textit{Reg-Net} had the same setting as that for $\mathcal{G}_{\boldsymbol{\Psi}}$, except being independently optimized using registration similarity $\mathcal{L}_{reg}$ and regularization $\mathcal{L}_{def}$ terms \citep{balakrishnan2019voxelmorph}. We ensured that the training dataset for the \textit{Seg-Net} and \textit{Reg-Net} was the same as that used for the Segis-Net framework. 

\subsection{Related methods involving segmentation and registration}
\label{ss:3.6}
When it comes to the combination of segmentation and registration, there are various integration strategies (Section \ref{S:1}). To investigate the benefit of the proposed simultaneous optimization strategy, we additionally compared Segis-Net with two previously published methods:
\begin{itemize}
	\item U-ReSNet for simultaneous segmentation and registration that used a shared feature encoder and separate decoders \citep{estienne2019u}.
	\item VoxelMorph for image registration alone that used correspondence in existing segmentation labels to boost registration \citep{balakrishnan2019voxelmorph}.
\end{itemize}

\subsection{Implementation} 
\label{ss:3.7}
For this diffusion MRI application, the segmentation ($\mathcal{F}_{\boldsymbol{\Theta}}$) and registration ($\mathcal{G}_{\boldsymbol{\Psi}}$) components of Segis-Net used different input images. Specifically, segmentation was based on the diffusion tensor image, as it contains directional information of fiber populations and was shown to be optimal in the present setting of clinical-quality resolution \citep{li2020neuro4neuro}. For spatial alignment, we adopted the input the commonly used scalar-value FA map derived from diffusion tensor imaging. 

To mitigate class imbalance and to improve computational efficiency, we combined the reference segmentation for the six tracts (Section \ref{ss:3.3}) into a three-channel map for using it as the segmentation ground truth $S$. This combination was possible since only few crossing fibers are expected between codirectional WM tracts (e.g., FMI and FMA). To evaluate performance on individual tracts after training, we extracted the two largest components from each of the three channels of the probabilistic prediction and subsequently identified the left and right (for CGC and CGH) or the anterior and posterior (for FMI and FMA) tract based on coordinates. 

The experiments of model training and evaluation were performed on an NVIDIA 1080Ti GPU and an AMD 1920X CPU. CNN-based methods were implemented using Keras-2.2.0 with a Tensorflow-1.4.0 backend and the Adam optimizer \citep{kingma2014adam}. For \textit{Reg-Net}, \textit{Seg-Net} and Segis-Net, weights of convolution kernels were initialized with the Glorot uniform distribution \citep{glorot2010understanding}. In each training epoch, input images were fed in random batches (size $=1$). Loss function hyperparameters were optimized based on segmentation and registration performance on the validation dataset (search range: [$10^{-3}, 10^{-2}, 10^{-1}, 10^{0}, 10^{1}, 10^{2}, 10^{3}$]); we set to $\alpha=10$, $\beta=0.01*\alpha$ for \textit{Reg-Net}; for Segis-Net we linearly increased $\alpha$ from $10$ to $100$ by $4$ per epoch (with $\beta$ changed accordingly), and set the additional parameter $\gamma=1$. The initial learning rates were experimentally optimized on the validation dataset and set to $1e^{-4}$, $1e^{-3}$ and $1e^{-3}$ for \textit{Reg-Net}, \textit{Seg-Net} and Segis-Net, which were decayed with a factor of 0.8 if the validation loss stopped decreasing for 10 epochs (decay condition, Algorithm \ref{algorithm1}). We stopped the training procedure at the point that the validation loss showed consecutive increases, i.e., early stopping \citep{bishop2006pattern}. The parameters of the model with the smallest error with respect to the validation dataset were used.

For VoxelMorph, the implementation as detailed by \cite{balakrishnan2019voxelmorph} was used directly. For U-ResNet, in contrast to the other tensor-based segmentation methods, we used the FA map as input for both segmentation and registration since the shared feature-encoder required the same input for both tasks. Affine registration was applied as a pre-processing step. Hyperparameters were tuned on the validation dataset; and we obtained improved performance by using an initial learning rate of $0.0005$ and by clipping the warped segmentation predictions into the range of [$10^{-7}$, $1-10^{-7}$].

\section{Experiments and results} \label{S:4} 

We applied the methods to analyze six WM tracts. The performance of the proposed Segis-Net was compared with the two baseline multi-stage pipelines on segmentation accuracy, registration accuracy, spatio-temporal consistency of segmentation, reproducibility of segmentation and measurements, and sample-size reduction; and compared with the two related methods in terms of the segmentation and registration accuracy.

\subsection{Segmentation accuracy} \label{ss:4.1}
Segmentation accuracy was quantified with respect to the reference segmentation (section~\ref{ss:3.4}) using the Dice coefficient metric. 

\begin{figure}[!ht]
\centering
\includegraphics[scale=0.46]{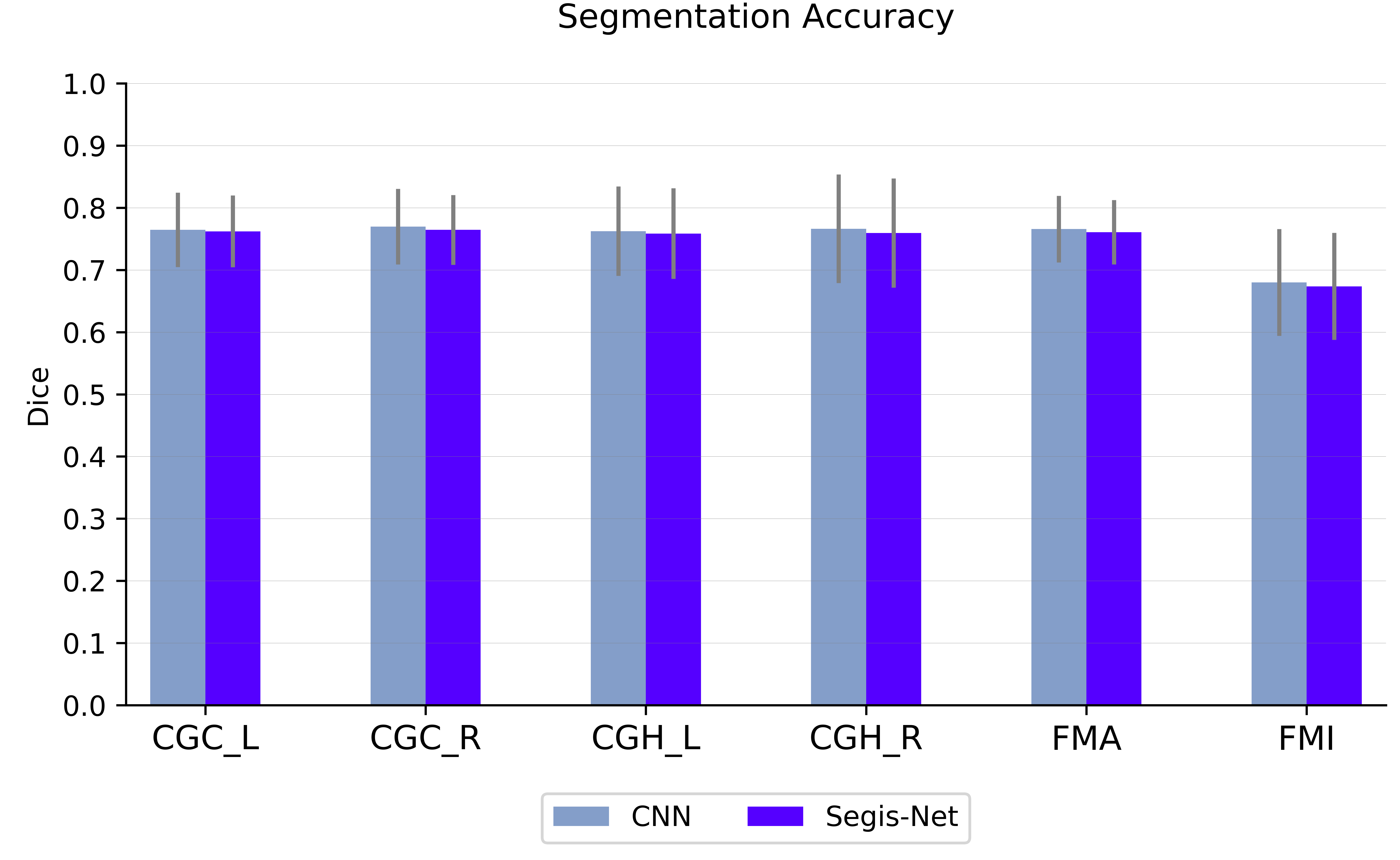}
\caption{Segmentation accuracies of the \textit{CNN} pipeline and Segis-Net for different tracts. Error bars indicate standard deviations.}\label{fig2}
\end{figure}

The proposed method yielded similar segmentation accuracy as the baseline multistage \textit{CNN} pipeline (\textit{Seg-Net}) for all six tracts (Figure \ref{fig2}). Both methods achieved relatively high accuracy in segmenting cingulum, i.e., the accuracy of left and right CGC and CGH tracts was around $0.76 \pm 0.07$. The accuracy was lowest for FMI (\textit{CNN}: $0.68 \pm 0.09$; Segis-Net: $0.67 \pm 0.09$), which is a thin and arch-shaped tract that is known to be more difficult to segment. Correcting for 6 tests resulted in an adjusted P-value threshold of $8.3 \times 10^{-3}$. There was no significant differences in segmentation accuracy between two methods.

The proposed method showed higher segmentation accuracy than U-ReSNet for all six tracts with a margin of around $10\%$ and smaller standard deviation (Table~\ref{tab1}).

\begin{table}[h!]
\centering
\small
 \begin{tabular}{ c|c|c } 
  \cline{1-3}
   & U-ReSNet & Segis-Net \\
   \cline{1-3} 
   CGC\_L & $0.69 \pm 0.06$ & \textbf{0.76} $\pm$ \textbf{0.06} \\ 
   CGC\_R & $0.69 \pm 0.07$ &  \textbf{0.76} $\pm$ \textbf{0.06} \\
   CGH\_L & $0.67 \pm 0.08$ &  \textbf{0.76} $\pm$ \textbf{0.07} \\
   CGH\_R & $0.67 \pm 0.09$ &  \textbf{0.76} $\pm$ \textbf{0.09} \\
   FMA    & $0.69 \pm 0.06$ & \textbf{0.76} $\pm$ \textbf{0.05} \\
   FMI    & $0.60 \pm 0.08$ & \textbf{0.67} $\pm$ \textbf{0.09} \\
   \cline{1-3}
 \end{tabular}
\caption{Segmentation Dice coefficient of U-ReSNet and Segis-Net. The bold value indicates a better performance in each row.}\label{tab1}
\end{table}

\subsection{Registration accuracy} \label{ss:4.2}
Registration accuracy of the approaches was evaluated with the spatial correlation (SC) similarity on the test dataset. According to the procedure in \cite{de2013improving}, the estimated transformation was applied to the continuous density maps of individual tracts obtained from probabilistic tractography, subsequently, the SC similarity between warped density maps was computed as follow:
\begin{equation}\label{eq12}
SC_k = \frac{\sum_{x \in \Omega_t} J_t^k  (J_s^k \circ \hat{\boldsymbol{\phi}})}{\bigg(\sum_{x \in \Omega_t}\sqrt{ (J^k_t)^2}\bigg) \, \bigg(\sum_{x \in \Omega_t}\sqrt{(J_s^k \circ \hat{\boldsymbol{\phi}})^2}\bigg)},
\end{equation}
where $J_t^k$ and $J_s^k$ indicate intensity of the target and source density image of the tract $k$. Despite a lot of intensity variation in the tract density maps across scans due to the probabilistic nature of tractography, higher intensity in general indicates more support for the tract while lower intensity conversely indicates increased uncertainty. Therefore, we assume that SC reflects the spatial correspondence of tracts.

\begin{figure}[!ht]
\centering
\includegraphics[scale=0.46]{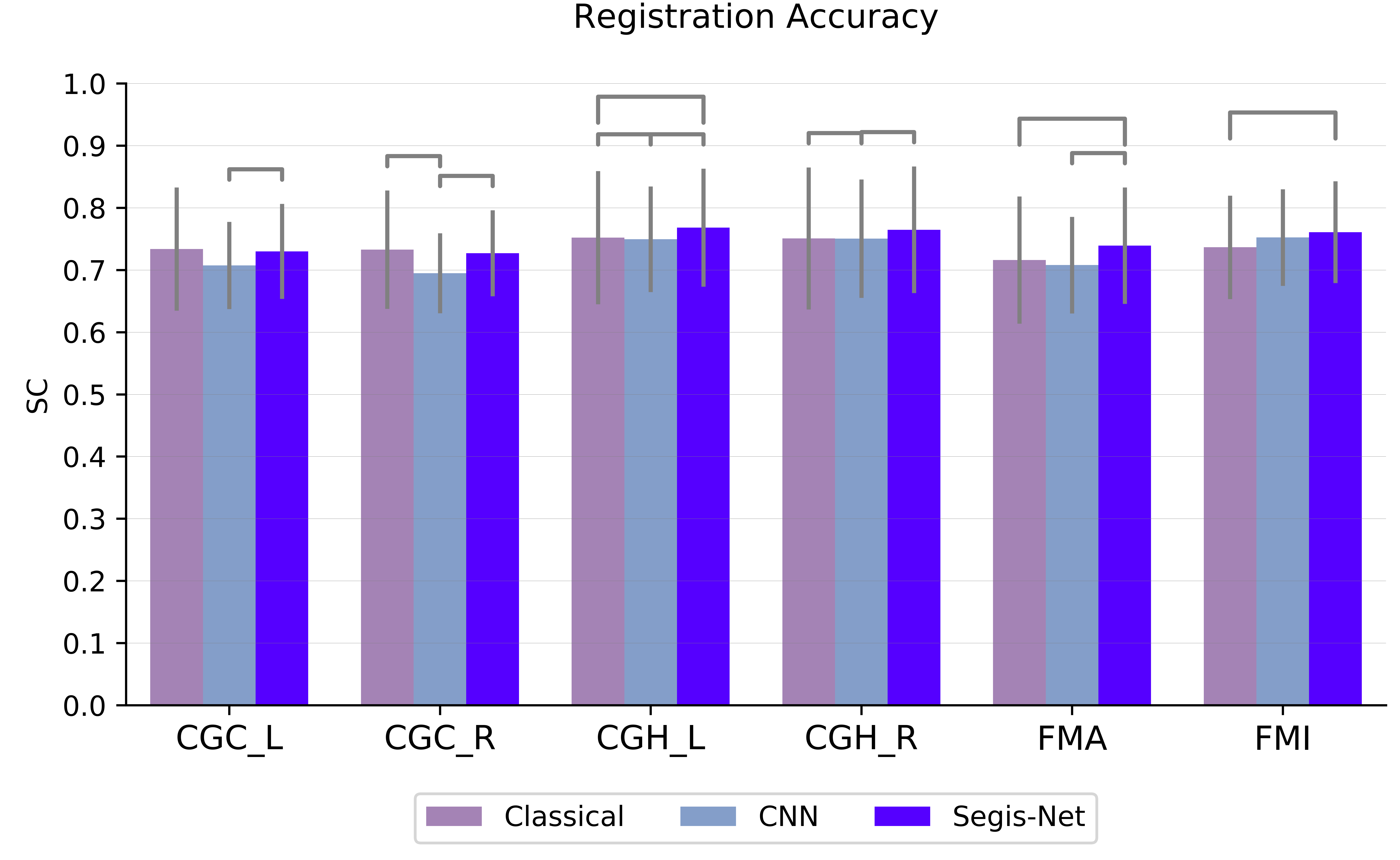}
\caption{Registration accuracies of the \textit{Classical}, \textit{CNN}, and Segis-Net pipeline as quantified by spatial correlation (SC) of the registered tract density maps. Error bars indicate standard deviations. The bracket hat indicates a significant difference between two methods (t-test, $p<2.8 \times 10^{-3}$).}\label{fig3}
\end{figure}

Figure \ref{fig3} presents the registration accuracy (SC) of Segis-Net and the baseline multistage pipelines. The SC in all six tracts was overall highest for the Segis-Net, followed by the \textit{Classical} pipeline. Correcting for 18 tests resulted in Bonferroni adjusted P-value threshold of $2.8 \times 10^{-3}$. Segis-Net results yielded a significantly better spatial correspondence than the \textit{Classical} pipeline in the left CGH (Segis-Net vs \textit{Classical} $= 0.77\pm 0.09$ vs $0.75\pm 0.11$), FMA ($0.74\pm0.09$ vs $0.72\pm0.10$), and FMI ($0.76\pm0.08$ vs $0.74\pm0.08$) tract. Statistically significant difference in the registration accuracy of Segis-Net and \textit{CNN} pipeline were observed in the left CGC (Segis-Net vs \textit{CNN} $= 0.73\pm0.08$ vs $0.71\pm0.07$), right CGC ($0.73\pm0.07$ vs $0.69\pm0.06$), left CGH ($0.77\pm0.09$ vs $0.75\pm0.08$), right CGH ($0.76\pm0.10$ vs $0.75\pm0.10$), and FMA ( $0.74\pm0.09$ vs $0.71\pm0.08$) tract. In general, the proposed Segis-Net approach achieved a better spatial correspondence than the two independently optimized registration algorithms using classical and learning-based techniques.

For comparing the proposed method with U-ReSNet and VoxelMorph, we added the performance metric that was used in their original papers \citep{estienne2019u,balakrishnan2019voxelmorph}, i.e., the Dice coefficient (DC) of registered reference segmentation of tracts, and added the common loss metric, i.e., the mean squared error (MSE) between registered FA maps (Table~\ref{tab2}). Generally, the SC similarity of the proposed method and U-ReSNet were better than that of VoxelMorph. Segis-Net led to the highest similarity in FMA and FMI tract; U-ReSNet was the highest for the left and right CGC, and the right of CGH tract; for the left CGH tract, a similar SC was observed for U-ReSNet and Segis-Net, although the variations were smaller in Segis-Net; Segis-Net achieved the best DC and MSE. For all six tracts, the DC of Segis-Net were higher than that of U-ReSNet, followed by VoxelMorph; the standard deviation of Segis-Net was overall smallest for all three metrics, except that of DC in three tracts (left and right CGH, and FMI) which were smallest for VoxelMorph.

\begin{table}[h!]
\centering
\small
 \begin{tabular}{ c|c|c|c|c } 
  \cline{1-5}
  \multicolumn{2}{c|}{} & U-ReSNet & VoxelMorph & Segis-Net \\
  \cline{1-5} 
  \multirow{6}{*}{SC} 
  & CGC\_L & \textbf{0.77} $\pm$ \textbf{0.11} & $0.72\pm0.09$ & $0.73\pm0.08$ \\ 
  & CGC\_R & \textbf{0.77} $\pm$ \textbf{0.11} & $0.71\pm0.09$ & $0.73\pm0.07$\\
  & CGH\_L & \textbf{0.77} $\pm$ \textbf{0.11} & $0.75\pm0.10$ & \textbf{0.77} $\pm$ \textbf{0.10} \\
  & CGH\_R & \textbf{0.77} $\pm$ \textbf{0.12} & $0.75 \pm 0.11$ & 0.76 $\pm$ 0.10 \\
  & FMA    & 0.73 $\pm$ 0.11 & $0.73\pm0.10$ & \textbf{0.74} $\pm$ \textbf{0.09}\\
  & FMI    & 0.75 $\pm$ 0.09 & $0.74\pm0.08$ & \textbf{0.76} $\pm$ \textbf{0.08}\\
  \cline{1-5}
  \multirow{6}{*}{DC} 
  & CGC\_L &$0.69\pm 0.07$  & $0.65\pm0.06$  & \textbf{0.74} $\pm$ \textbf{0.06} \\ 
  & CGC\_R & $0.70\pm 0.07$ & $0.65\pm0.06$ & \textbf{0.74} $\pm$ \textbf{0.05} \\
  & CGH\_L & $0.67\pm0.08$  & $0.64\pm0.07$ & \textbf{0.71} $\pm$ \textbf{0.08} \\
  & CGH\_R &$0.67\pm0.10$   & $0.64\pm 0.08$ & \textbf{0.71} $\pm$ \textbf{0.09} \\
  & FMA    & $0.70\pm0.06 $ & $0.68\pm 0.06$ & \textbf{0.72} $\pm$ \textbf{0.06} \\
  & FMI    &$0.57\pm 0.10$  & $0.56\pm 0.07$ & \textbf{0.60} $\pm$ \textbf{0.09} \\
  \cline{1-5}  
  \multicolumn{2}{c|}{MSE ($\times 10^{-2}$)} & $0.47 \pm 0.38$ & $0.19\pm0.88$ & \textbf{0.13} $\pm$ \textbf{0.10} \\
  \cline{1-5} 
 \end{tabular}
\caption{Registration performance of U-ReSNet, VoxelMorph and Segis-Net, as quantified by the spatial correlation (SC) similarity, the Dice coefficient (DC), and the mean squared error (MSE). The bold value indicates the best performance in each row.}\label{tab2}
\end{table}

\subsection{Spatio-temporal consistency of segmentation} \label{ss:4.3}
To evaluate the spatio-temporal consistency of segmentation (STCS) for Segis-Net and the baseline multistage pipelines, we measured the correspondence between warped segmentation results across time-points using the Dice coefficient. The consistency of each tract was averaged over two directions by reversing the target and source image, which for the tract $k$ can be formulated as: 
\begin{equation}\label{eq13}
    STCS_k = \frac{1}{2} \bigg(\frac{2|\hat{S}_t^k \cap \hat{S}_s^k \circ \hat{\boldsymbol{\phi}} |}{|\hat{S}_t^k| + |\hat{S}_s^k \circ \hat{\boldsymbol{\phi}|}} + \frac{2|\hat{S}_s^k \cap \hat{S}_t^k \circ \hat{\boldsymbol{\phi}^{-1}} |}{|\hat{S}_s^k| + |\hat{S}_t^k \circ \hat{\boldsymbol{\phi}^{-1}|}}\bigg).
\end{equation}

Each pipeline was evaluated as a whole, that is, 1) in \textit{Classical} pipeline the reference segmentation was warped by Elastix algorithm, 2) in \textit{CNN} pipeline the prediction of \textit{Seg-Net} was warped by the predicted transformation of the \textit{Reg-Net}, and 3) in Segis-Net framework the segmentation prediction in native space and the segmentation warped from another time-point were available after a bidirectional test. 

\begin{figure}[!ht]
\centering
\includegraphics[scale=0.46]{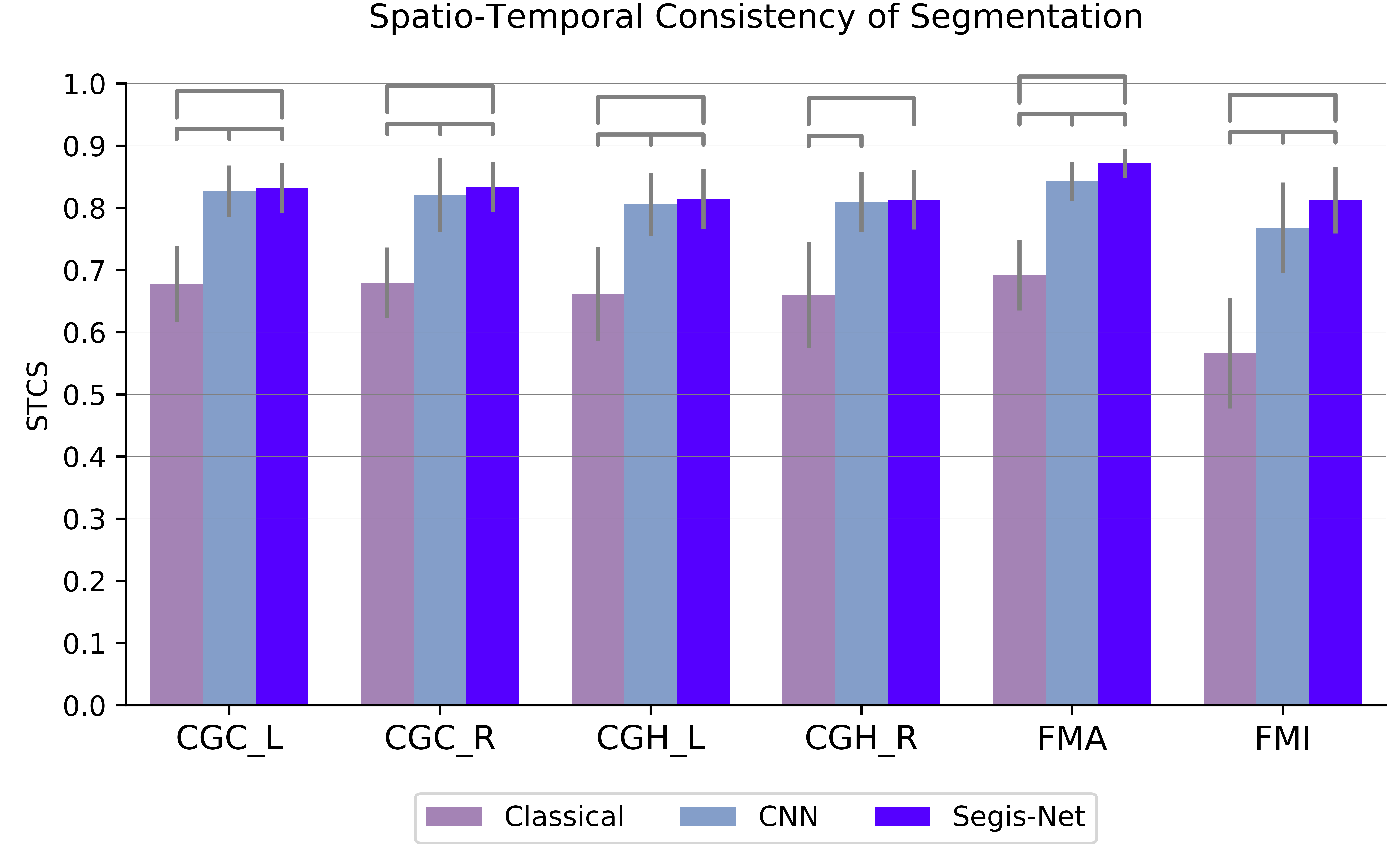}
\caption{Spatio-temporal consistency of segmentation (STCS) with the \textit{Classical}, \textit{CNN}, and Segis-Net pipeline. Error bars indicate standard deviations. The bracket hat indicates a significant difference between two methods (t-test, $p<2.8 \times 10^{-3}$).}\label{fig4}
\end{figure}

The proposed Segis-Net overall showed higher segmentation consistency than the \textit{CNN} and the \textit{Classical} pipeline (Figure \ref{fig4}). Correcting for 18 tests resulted in an adjusted P-value threshold of $2.8 \times 10^{-3}$. In comparison with the \textit{CNN} pipeline, Segis-Net results yielded significantly higher spatio-temporal consistency in left CGC (Segis-Net vs \textit{CNN} $= 0.83\pm0.04$ vs $0.82\pm0.04$), right CGC ($0.83\pm0.04$ vs $0.82\pm0.06$), left CGH ($0.82\pm0.05$ vs $0.81\pm0.05$), FMA ($0.87\pm0.02$ vs $0.84\pm0.03$), and FMI ($0.81\pm0.05$ vs $0.77\pm0.07$) tract. In all six tracts, Segis-Net significantly outperformed the \textit{Classical} pipeline, i.e., in left CGC (Segis-Net vs \textit{Classical} $=0.83\pm0.04$ vs $0.68\pm0.06$), right CGC ($0.83\pm0.04$ vs $0.68\pm0.06$), left CGH ($0.82\pm0.05$ vs $0.66\pm0.08$), right CGH ($0.81\pm0.05$ vs $0.66\pm0.09$), FMA ($0.87\pm0.02$ vs $0.69\pm0.06$), and FMI ($0.81\pm0.05$ vs $0.57\pm0.09$) tract.

\subsection{Reproducibility of segmentation and measurements} \label{ss:4.4}
Reproducibility of tract-specific segmentations, volumes, and diffusion metrics of the pipelines was evaluated using the reproducibility dataset. We quantified voxel-wise agreement between segmentations of repeated scans using Cohen's kappa coefficient ($\kappa$). The segmentations ($\hat{\mathcal{S}}_t\,, \hat{\mathcal{S}}_s$) were obtained in the native space, and subsequently aligned ($\hat{\mathcal{S}}_s \circ \hat{\boldsymbol{\phi}}$). Kappa $\kappa$ of the tract $k$ is defined as: 
\begin{equation}
\kappa_k =\frac{p_o(\hat{S}_t^k, \hat{S}_s^k \circ \hat{\boldsymbol{\phi}}) - p_e(\hat{S}_t^k, \hat{S}_s^k \circ \hat{\boldsymbol{\phi}})}{1-p_e(\hat{S}_t^k, \hat{S}_s^k \circ \hat{\boldsymbol{\phi}})}\label{eq14},
\end{equation}
in which $p_o(\hat{S}_t^k, \hat{S}_s^k \circ \hat{\boldsymbol{\phi}})$ is the observed agreement between $\hat{S}_t^k$ and $\hat{S}_s^k \circ \hat{\boldsymbol{\phi}}$\,, $p_e$ is the hypothetical probability of the agreement. Given $|\Omega_t|$ being the total number of voxels in the target image, $|S|$ and $|\Omega_t|-|S|$ being the number of tract and non-tract voxels, the observed agreement (i.e., accuracy) is computed as:
\begin{equation}\label{eq15}
p_o(\hat{S}_t^k, \hat{S}_s^k \circ \hat{\boldsymbol{\phi}})=\frac{|\hat{S}_t^k \cap (\hat{S}_s^k \circ \hat{\boldsymbol{\phi}})| + |(1-\hat{S}_t^k) \cap \big(1-(\hat{S}_s^k \circ \hat{\boldsymbol{\phi}})\big) |}{|\Omega_t|},
\end{equation}
the hypothetical probability of the agreement can be formulated as:
\begin{equation}\label{eq16}
p_e(\hat{S}_t^k, \hat{S}_s^k \circ \hat{\boldsymbol{\phi}})=\frac{1}{|\Omega_t|^2} \big(|\hat{S}_t^k| \times |\hat{S}_s^k \circ \hat{\boldsymbol{\phi}} | + (|\Omega_t| - |\hat{S}_t^k|) \times (|\Omega_t| - |\hat{S}_s^k \circ \hat{\boldsymbol{\phi}} |)\big).
\end{equation}
Typically, a $\kappa > 0.60$ indicates ``substantial'' agreement, and a $\kappa > 0.80$ indicates ``almost perfect'' agreement \citep{landis1977application}. 

\begin{figure*}[!ht]
\centering
     \subfigure[]{
     \includegraphics[scale=0.46]{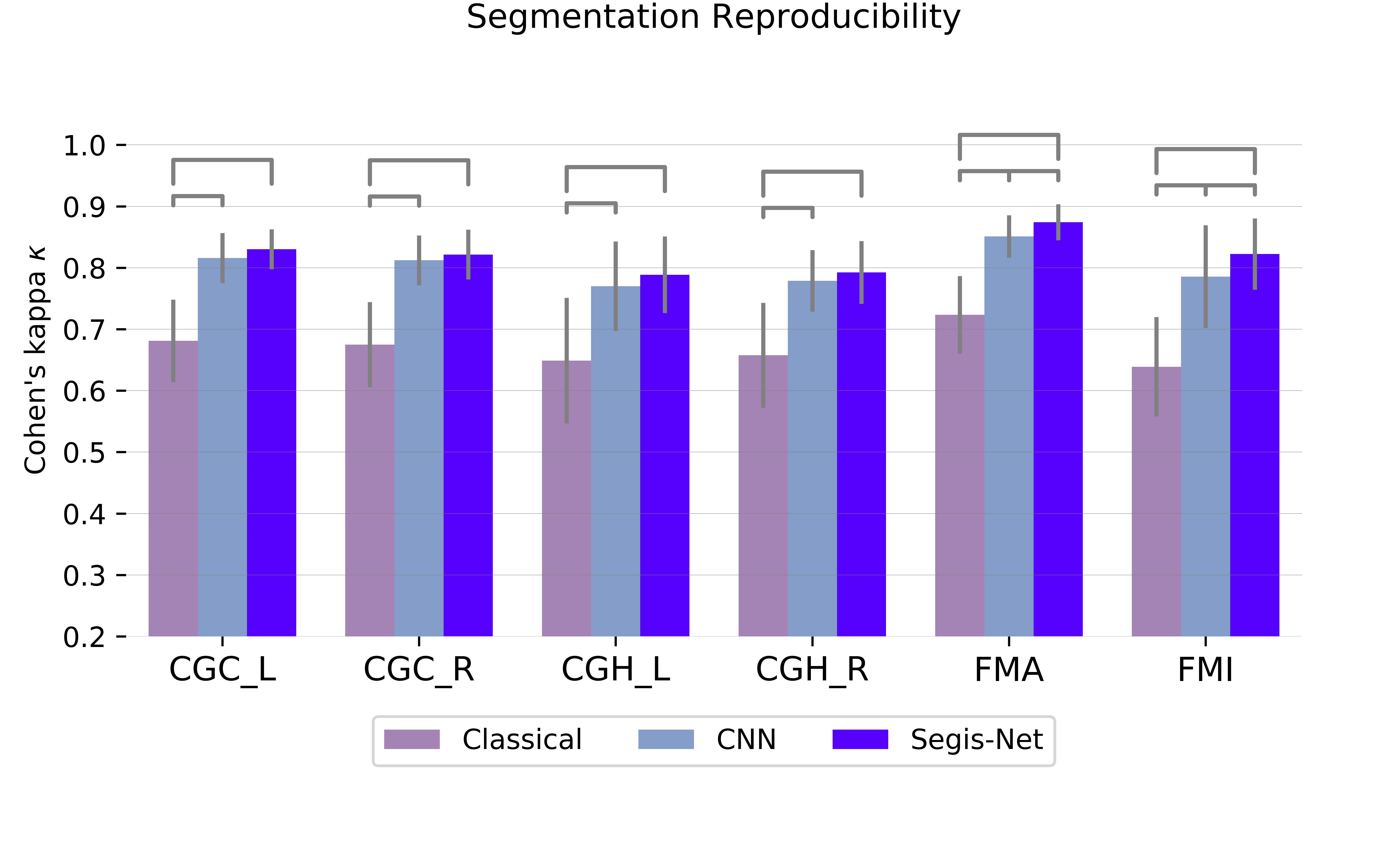}}
     \quad
     \subfigure[]{
     \includegraphics[scale=0.46]{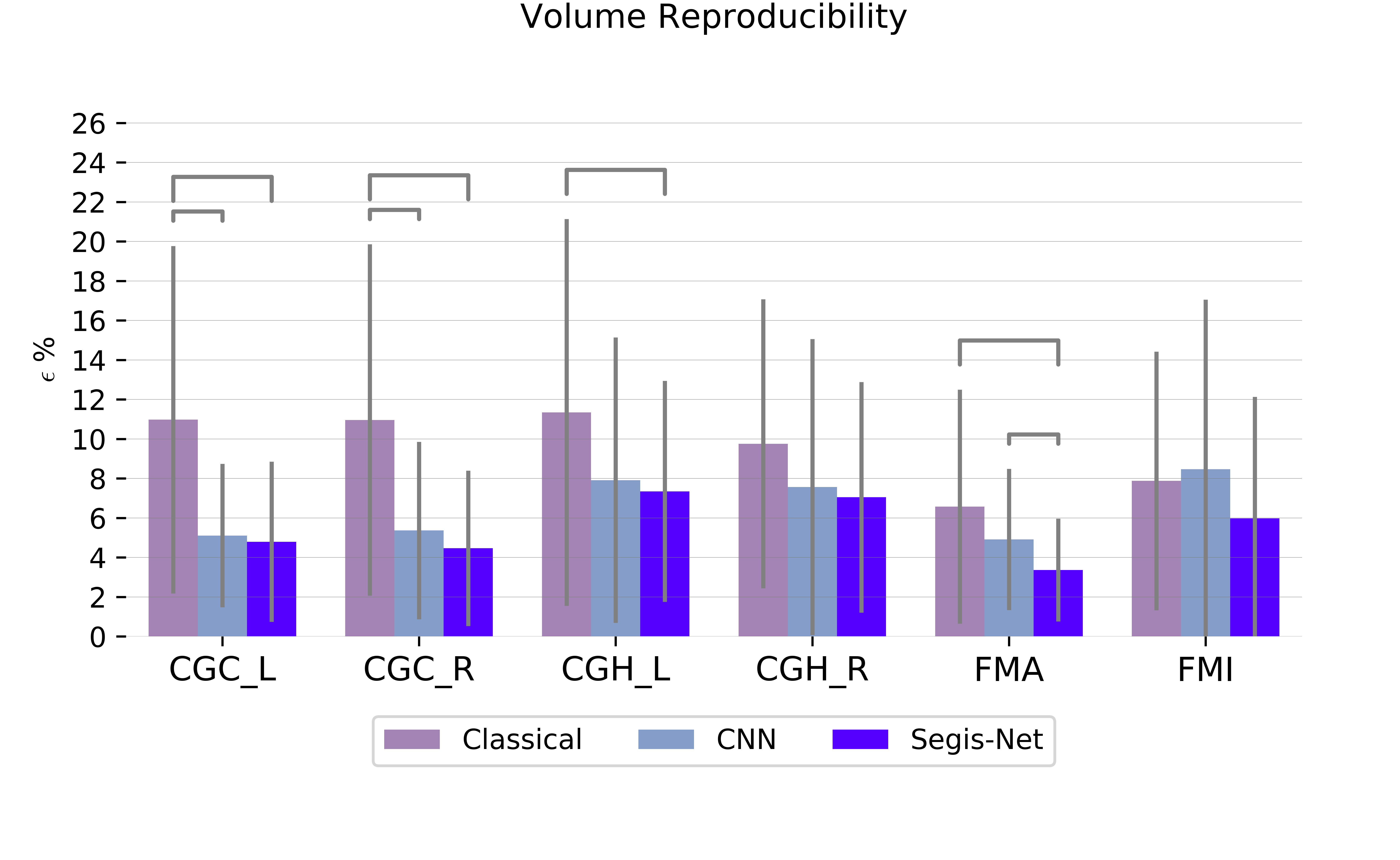}} 
     
     \subfigure[]{
  	 \includegraphics[scale=0.46]{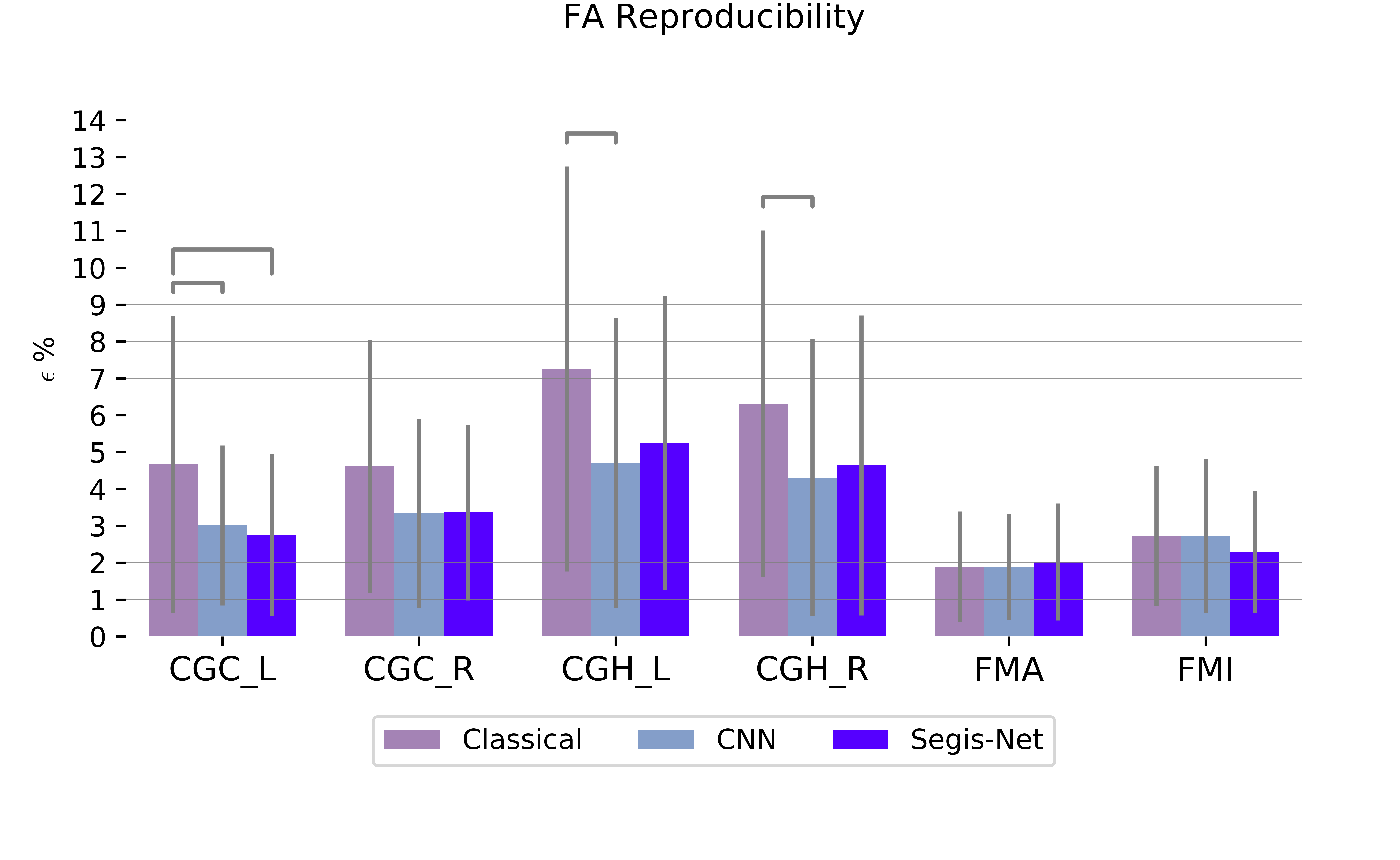}}
  	 \quad
     \subfigure[]{
     \includegraphics[scale=0.46]{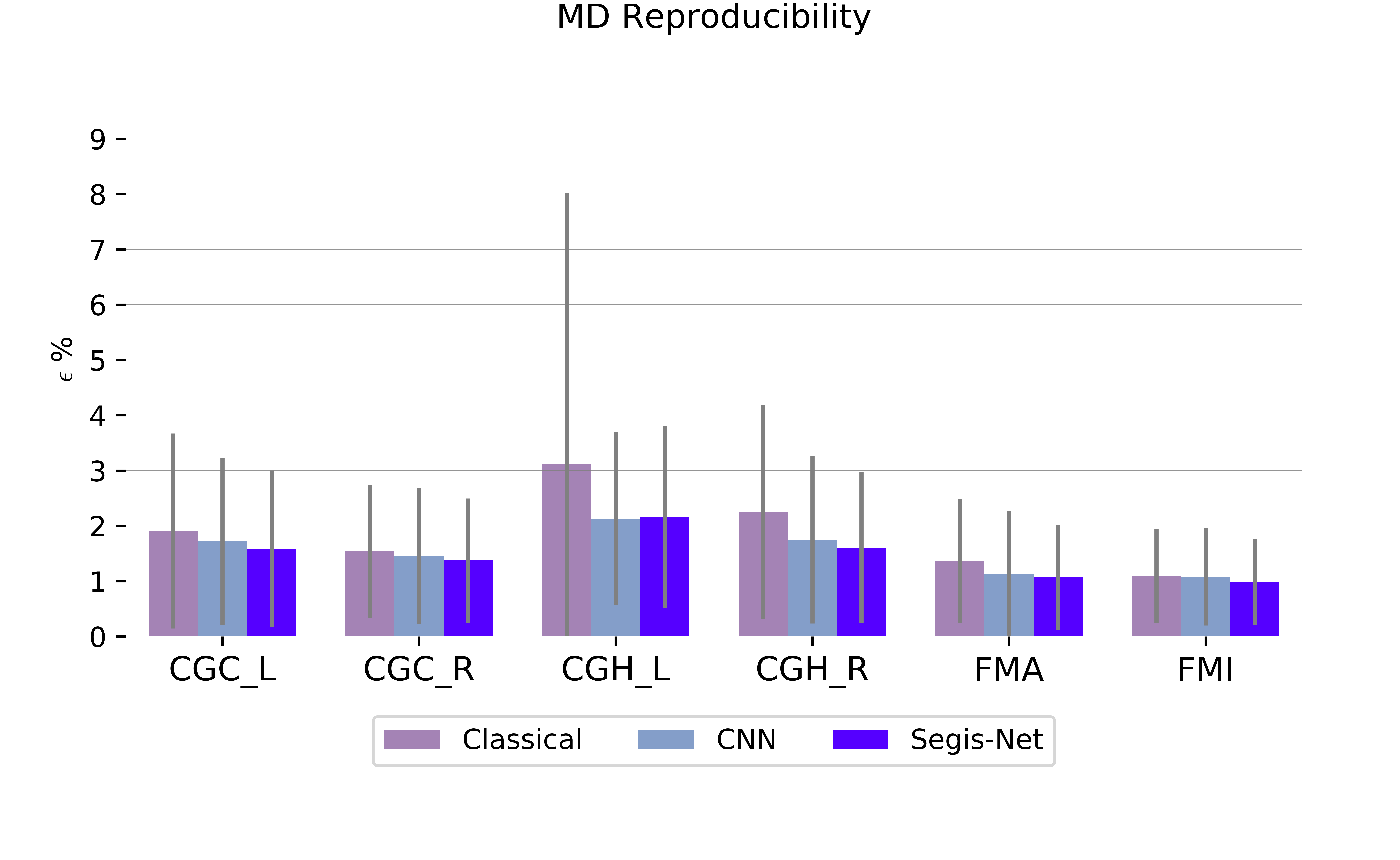}}     
\caption{Reproducibility of tract-specific measures with the \textit{Classical}, \textit{CNN}, and Segis-Net pipeline. Error bars indicate standard deviations. The bracket hat indicates a significant difference between two methods (t-test, $p<2.8 \times 10^{-3}$). In figure (a), a higher Cohen's kappa coefficient ($\kappa$) indicates a better reproducibility. In figure (b-d), a lower error ($\epsilon \%$) indicates a better reproducibility. Volume: tract-specific volume (ml), FA: fractional anisotropy, MD: mean diffusivity ($10^{-3}mm^2/s$).}\label{fig5}
\end{figure*}

Similarly, to evaluate the reproducibility of tract-specific measurements, we computed the FA, MD and volume in image native space, and subsequently assessed relative differences in paired scan-rescan measures ($m_t\,, m_s$) as an indicator of measurement error ($\epsilon$), i.e.,
\begin{equation}\label{eq17}
\epsilon =\frac{ 2\lvert m_{s} - m_{t} \lvert }{(m_{s} + m_{t})} \times 100 \%.
\end{equation}
For FA and MD, the tract-specific measures were quantified as the median of non-zero values within the segmented images. A lower $\epsilon$ indicates a better reproducibility. 

Figure \ref{fig5} presents the reproducibility of tract-specific segmentation and measures determined with the baseline multi-stage pipelines and Segis-Net. The proposed Segis-Net achieved the best segmentation reproducibility, followed by the \textit{CNN} pipeline (Figure \ref{fig5} (a)); in all six tracts, $\kappa$ was around $0.80$ or higher, indicating ``almost perfect'' agreements between segmentations of repeated scans. Correcting for 18 tests for each metric resulted in an adjusted P-value threshold of $2.8 \times 10^{-3}$, resulting in overall statistically significant improvement by Segis-Net over the \textit{Classical} pipeline. For two tracts, voxel-wise agreement of Segis-Net was significantly higher than that of the \textit{CNN} pipeline, i.e., FMA (Segis-Net vs \textit{CNN} $= 0.87\pm0.03$ vs $0.85\pm0.03$) and FMI ($0.82\pm0.06$ vs $0.79\pm0.08$). 

Additionally, in the evaluation of the reproducibility in tract-specific volume measures, Segis-Net showed the smallest error in all six tracts (Figure \ref{fig5} (b)). The error of Segis-Net was significantly smaller than the \textit{Classical} pipeline in left CGC (Segis-Net vs \textit{Classical} $=4.8\pm4.1\%$ vs $11\pm8.8 \%$), right CGC ($4.5\pm3.9\%$ vs $11\pm8.9 \%$), left CGH ($7.3\pm5.6\%$ vs $11\pm9.8\%$), and FMA ($3.4\pm2.6\%$ vs $6.6\pm5.9\%$) tract. This outperformed the \textit{CNN} pipeline significantly in the FMA tract (Segis-Net vs \textit{CNN} $=3.4\pm2.6\%$ vs $4.9\pm3.6 \%$). Reproducibility of FA and MD measurements was similar for the three methods (Figure \ref{fig5} (c, d)). For the CGH and left CGC tracts, the reproducibility of FA using the \textit{CNN} pipeline was significantly higher than that of the \textit{Classical} pipeline. Segis-Net outperformed the FA reproducibility of the \textit{Classical} pipeline only in the left CGC tract. For MD, no significant improvement over the \textit{Classical} pipeline was observed. A table (Table~\ref{tab3}) with the results of Figure 3-6 is provided in the supplementary files.

\subsection{Sample-size reduction} \label{ss:4.5}
An implication of the reduced measurement error ($\epsilon$) is that fewer participants or time-points would be required to achieve the same statistical power, i.e., a smaller sample size. We followed \cite{diggle2002analysis} and \cite{reuter2012within} to estimate the percentage of the sample sizes ($P$) that would be required for each of the pipelines:

\begin{equation}\label{eq18}
P_{ij} = \frac{\sigma_i^2 \times (1-\rho_i)}{\sigma_j^2 \times (1-\rho_j)} \times 100 \%,
\end{equation}
where $\sigma_i$ and $\sigma_j$ are standard deviations in the measurements determined with the pipeline $i$ and $j$, and $\rho_i$ and $\rho_j$ are the correlation coefficients between the repeated measurements determined with the two pipelines.

\begin{figure*}[!ht]
\centering
\subfigure[Volume Measures]{
    \includegraphics[width=0.32\textwidth]{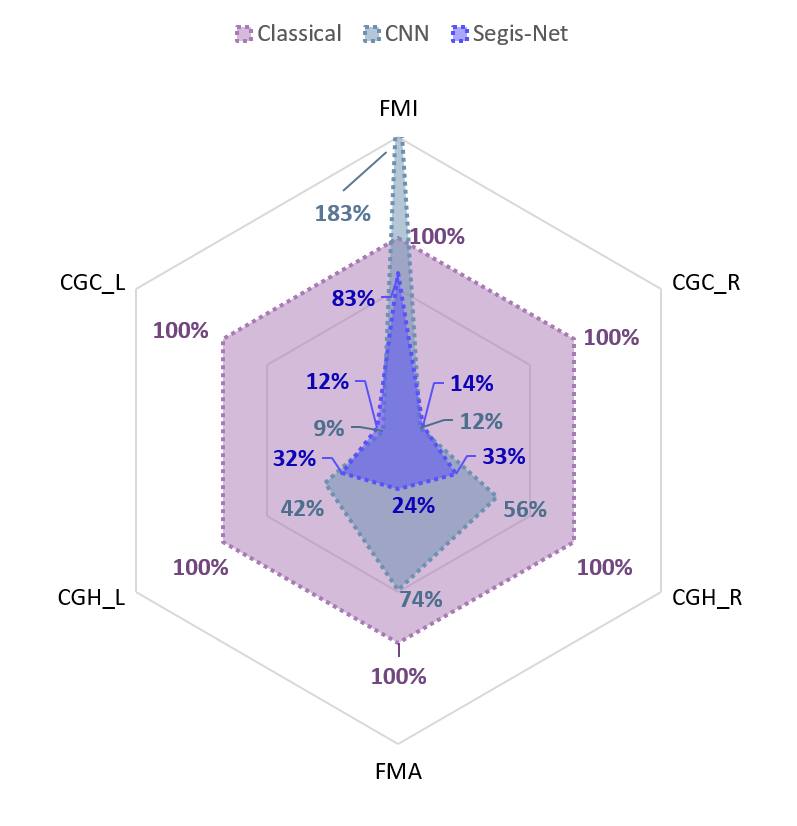}}
\subfigure[FA Measures]{
     \includegraphics[width=0.32\textwidth]{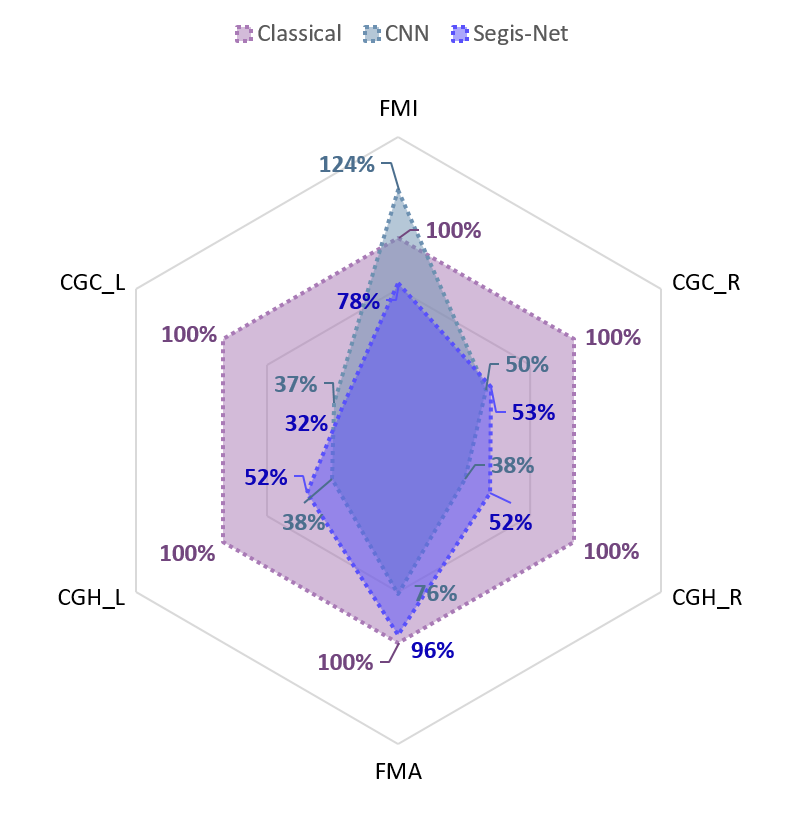}}
\subfigure[MD Measures]{
     \includegraphics[width=0.32\textwidth]{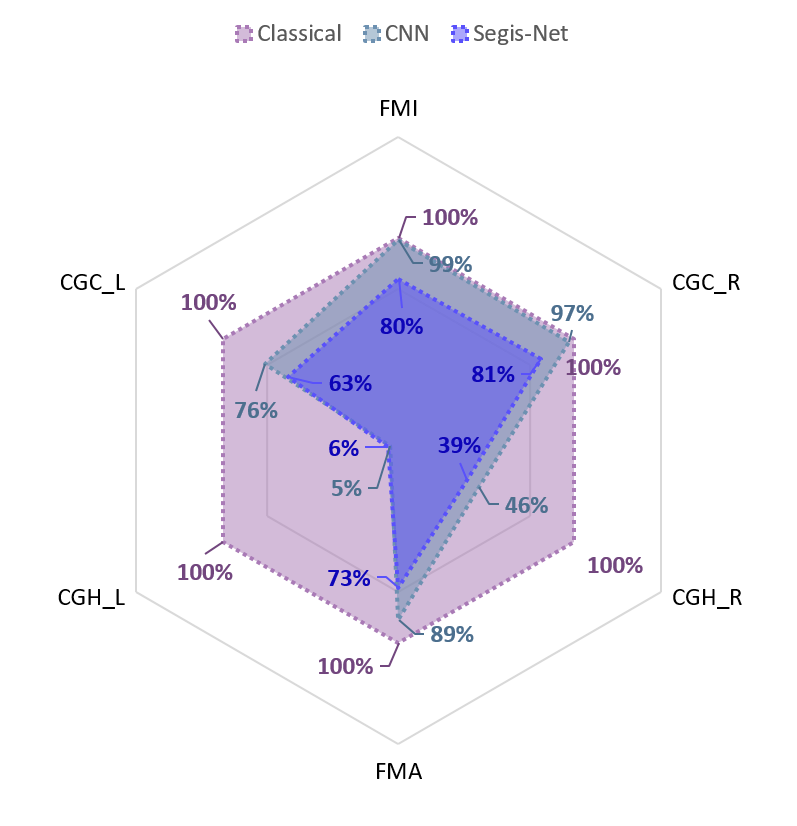}}

\caption{The percentage of sample-size that would be required in tract measures of volume, FA, and MD with the \textit{CNN} pipeline and Segis-Net. The sample-size required for the \textit{Classical} pipeline is used as the reference (100\%).}\label{fig6}
\end{figure*}

Figure \ref{fig6} presents the percentage of sample-size reduction that could be achieved by the \textit{CNN} and the proposed Segis-Net compared to the \textit{Classical} pipeline. In line with the reproducibility results, the data analyzed with Segis-Net would overall require the least sample-size to achieve the same statistical power. The percentage of reduction was especially remarkable in volume measures, in which on average only $33.0\%$ of data would be required. The average percentage of reduction was $60.5\%$ for FA and $57.0\%$ for MD. Several percentages of the \textit{CNN} pipeline were smaller than those of the Segis-Net, e.g., in FA measures of CGH and FMA tract (Figure \ref{fig6} (b)), but its performance showed to be less stable across tracts than the Segis-Net, which in all settings consistently decreased in the required samples over the \textit{Classical} pipeline. The percentage of reduction was generally similar for left/right homologous tracts except for the MD measure in the left of CGH (Figure \ref{fig6} (c)). This large reduction could be related to the MD reproducibility of the \textit{Classical} pipeline, in which the left CGH tract had a much higher variation in errors comparing with that of the other tracts (Figure \ref{fig5} (d)).

\section{Discussion}
\label{s:5}
We developed a single-step deep learning framework, coined Segis-Net, for simultaneous optimization of segmentation and registration. The method was applied to analyze changes in WM tracts from a large set of longitudinal diffusion MRI images. To evaluate the performance of the method, we compared it with two state-of-the-art methods, and two multistage pipelines consisting of independent segmentation and registration components, i.e., the \textit{Classical} and \textit{CNN} pipeline. Segis-Net advanced the state-of-the-art by a higher segmentation and registration accuracy, and led to improved performances in registration accuracy, spatio-temporal consistency of segmentation, and reproducibility of segmentation and tract-specific measures comparing with the multi-stage pipelines. We evaluated the practical value of the improved performance in terms of sample-size reduction that could be achieved when employing the method. The tract-specific mesures analyzed with Segis-Net would only require $33.0\% - 60.5\%$ sample-size of the data for achieving the same effect size as the \textit{Classical} pipeline. 

To date most developments in longitudinal analysis frameworks have focused on unbiased ways of registering image time series \citep{metz2011nonrigid,keihaninejad2013unbiased}, in which a multistage approach combing independent segmentation and registration components is often used \citep{de2013changes,yendiki2016joint}. In this paper, we aimed to investigate a different way to improve the performance of the longitudinal framework by using a single-step CNN that optimizes both tasks simultaneously. The sole value of simultaneous optimization was demonstrated by the comparison with the \textit{CNN} pipeline. There was no benefit observed for segmentation alone, but for registration, spatio-temporal consistency of segmentation, and reproducibility, simultaneous optimization led to significantly improved performance.  

In the evaluation of segmentation performance, similar accuracies for the \textit{CNN} and Segis-Net framework was observed for the six tracts (Figure \ref{fig2}). Relative segmentation accuracy between individual tracts were in line with those reported in literature \citep{wasserthal2018tractseg,li2020neuro4neuro}. For instance, a small and curved object like the FMI tract tended to have a lower Dice coefficient than the larger FMA tracts. For all six tracts, the proposed Segis-Net showed a better segmentation performance than U-ReSNet, an existing simultaneous method (Table~\ref{tab1}). We expect the added value of Segis-Net to be related to two factors: 1) the method allows the use of diffusion tensor images for tract segmentation, as we use parallel network modules and only align the predicted segmentation; this circumvents the need to interpolate tensor images. In other words, task-specific inputs can be used; and 2) the sub-branches in the segmentation stream (Figure~\ref{fig:s1}) are designed for the prediction of white matter tracts which can overlap with each other, unlike the exclusive tissue labels focused by other works.

In the task of registration alone, Segis-Net overall yielded the best accuracy among the methods. It significantly outperformed the \textit{Classical} pipeline for three tracts and the \textit{CNN} pipeline for five tracts (Figure \ref{fig3}). This is an important observation as 1) it showed that simultaneous optimization was beneficial to one of the individual tasks, and 2) it is non-trivial to improve registration accuracy over a classical algorithm, in which the transformation is pair-wise optimized on the test images. During the comparison with the state-of-the-art methods, we observed two interesting results (Table~\ref{tab2}). First, VoxelMorph was the only method that directly optimized on the DC metric, but it led to a lowest DC score. This can be due to the fact that the segmentation labels used in diffusion imaging studies are often independently obtained for each image, which is much less correlated to the registration performance than is the case for atlas-based segmentation \citep{balakrishnan2019voxelmorph}. As a result, the alignment of ``imperfect'' segmentation labels can be an obstructive loss term instead. Second, although the MSE of U-ReSNet was almost four times that of the Segis-Net, it achieved a good SC similarity, especially in the thin structures (CGC and CGH). This can be attributed to the formulation of their registration loss as the sum of local cross-correlation and MSE.

In all six tracts, we observed substantially higher spatio-temporal consistency of segmentation and reproducibility of segmentation with Segis-Net than with the two multistage pipelines (Figure \ref{fig4}, \ref{fig5}). The spatio-temporal consistency of segmentation as quantified by the Dice coefficient ranged $0.81 - 0.87$ for Segis-Net, significantly outperforming the \textit{Classical} pipeline for all the six tracts (range: $0.57-0.69$) and the \textit{CNN} pipeline for five tracts (range: $0.77-0.84$). The segmentation reproducibility as quantified by Cohen's kappa ranged $0.79 - 0.87$ for Segis-Net, significantly higher than the \textit{Classical} pipeline for all the six tracts (range: $0.64-0.72$) and the \textit{CNN} pipeline for two tracts (range: $0.77-0.85$). These results indicate that Segis-Net can serve as a reliable alternative to the \textit{Classical} pipeline in spatially capturing macro-structural brain changes over time.  

In addition, more significant improvements were observed for the reproducibility of tract-specific volume assessment, but not for the FA and MD measures. For volume reproducibility, Segis-Net yielded the least error in the measurements of scan and re-scan, followed by the \textit{CNN} pipeline (Figure \ref{fig5}). For the FA and MD measures, we observed relatively similar reproducibility for the three methods, in which significant difference was only observed in FA reproducibility of CGH and left CGC tract. This suggests that diffusion measures are quite robust to variations in the geometry of the segmented tract. It's worth noting that the FA reproducibility of the \textit{Classical} pipeline could be higher than the benchmark of tractography-based segmentation methods, since it is optimized on the FA reproducibility on a subset of the data.

These improved performances have practical values in a power analysis, where both the \textit{CNN} pipeline and Segis-Net showed to be able to reduce the required sample-size to achieve the same statistical power as the \textit{Classical} pipeline. The data processed with Segis-Net would require on average $33.0\%$ of the sample-size for volume measures, $60.5\%$ for FA, and $57.0\%$ for MD measures, requiring consistently a decreased sample-size for all the settings. The averaged percentages for the \textit{CNN} pipeline were $62.7\%$, $60.5\%$ and $68.7\%$. For FMI tract, it would, however, require $183\%$ and $124\%$ of the sample-size for the volume and FA measures. The observed dispersion of sample-size reduction with the \textit{CNN} pipeline may suggest that simultaneous optimization was beneficial to the robustness of the method across the concurrently segmented tracts. 

Whereas the method is generic, we specifically implemented and optimized it for longitudinal study in diffusion MRI data. In diffusion MRI application, we adopted the commonly used scalar-value FA map as the input for registration. Deformable registration of diffusion tensor images is known to be challenging due to the directional components contained in voxels. Despite developments in classical methods for tensor reorientation during the optimization \citep{cao2006diffeomorphic,zhang2007high}, for learning-based registration it still largely remains unexplored. With the promising results of diffusion tensor interpolation as shown by \cite{grigorescu2020diffusion}, Segis-Net based on solely tensor images would be an interesting direction to explore.

The Segis-Net framework presented in the current study is limited to two time-points. This is because learning-based registration algorithms currently only support pairwise transformations \citep{balakrishnan2019voxelmorph}. One limitation of our method is therefore that it does not allow for analysis of arbitrary number of time-points. In the present study, we grouped the available triple time-points from the same participant into orderless image-pairs for bidirectional analysis. A future possible improvement of the method could be extending the registration component of Segis-Net to enable learning-based group-wise analysis of a set of time-points \citep{li2020learning}. 

Beyond the current application, we expect that this work could be extended to other imaging sequences and for example for segmentation of lesion images. For future work, we plan to adapt the proposed method to analyze brain diseases with large and progressive changes. For instance, registration of brains with lesions due to cortical infarct may benefit from a simultaneous segmentation of infarct regions.

\section{Conclusion}
\label{s:6}

We proposed a single-step deep learning framework for longitudinal diffusion MRI analysis, in which segmentation and deformable registration were integrated for simultaneous optimization. The comparison with baseline multistage approaches and state-of-the-art methods showed that the proposed Segis-Net can be applied as a reliable tool to support spatio-temporal analysis of WM tracts from longitudinal diffusion MRI imaging. Besides the improved performances, a two-in-one framework for concurrent segmentation and registration also enables a light-weight way of fast quantification of brain changes overtime. This may lead to a more prominent role for tract-specific biomarkers in applications where tract segmentation and registration are subject to time constraints. With the increasing availability of longitudinal diffusion data, we expect future studies investigating progressive neurodegeneration can greatly benefit from the improved reliability and efficiency of Segis-Net.

\section*{Acknowledgments}
This work was sponsored through grants of the Medical Delta Diagnostics 3.0: Dementia and Stroke, the EU Horizon 2020 project EuroPOND (666992), the Netherlands CardioVascular Research Initiative (Heart-Brain Connection: CVON2012-06, CVON2018-28), and the Dutch Heart Foundation (PPP Allowance, 2018B011).

\section*{Data availability}
The datasets analyzed during the current study are not publicly available. Due to the sensitive nature of the data used in this study, participants were assured raw data would remain confidential and would not be shared.

\section*{Code availability}
The code for Segis-Net, the \textit{CNN} pipeline, as well as the implementation of Elastix is available at 
\url{https://gitlab.com/blibli/segis-net}. 

\section*{Ethics statement}
The Rotterdam Study has been approved by the Medical Ethics Committee of the Erasmus MC (registration number MEC 02.1015) and by the Dutch Ministry of Health, Welfare and Sport (Population Screening Act WBO, license number 1071272-159521-PG). All participants provided written informed consent to participate in the study and to have their information obtained from treating physicians \citep{ikram2020objectives}.

\bibliographystyle{model2-names.bst}\biboptions{authoryear}
\bibliography{hybrid_neuroimage.bib}

\section*{Supplementary Material}
\begin{figure*}[!ht]
\centering
\includegraphics[width=0.95\textwidth]{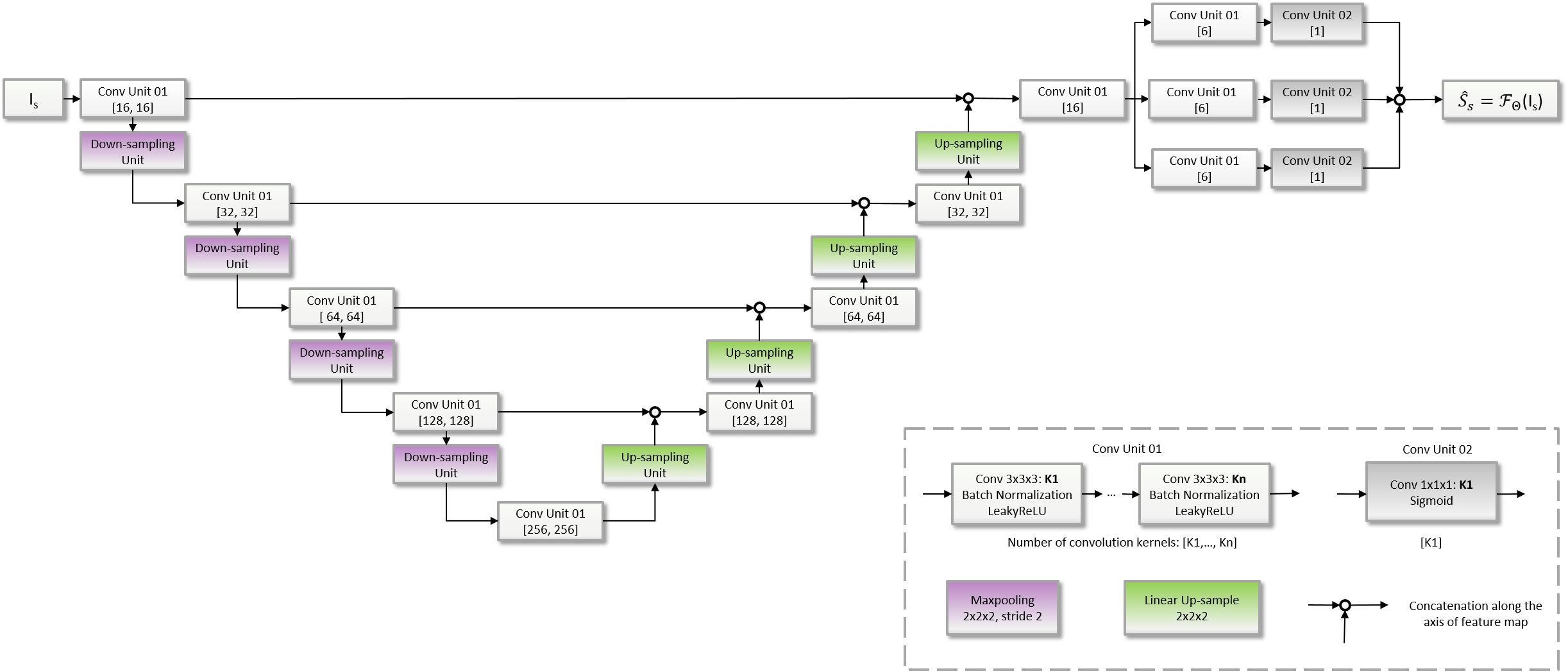}
\caption{The network architectures of $\mathcal{F}_{\Theta}$ module used in the present study. The number of convolution kernels used in Conv Unit 01 is denoted as [K1, ..., Kn], and that for Conv Unit 02 as [K1]}.\label{fig:s1}
\end{figure*}

\begin{figure*}[!ht]
\centering
\includegraphics[width=0.95\textwidth]{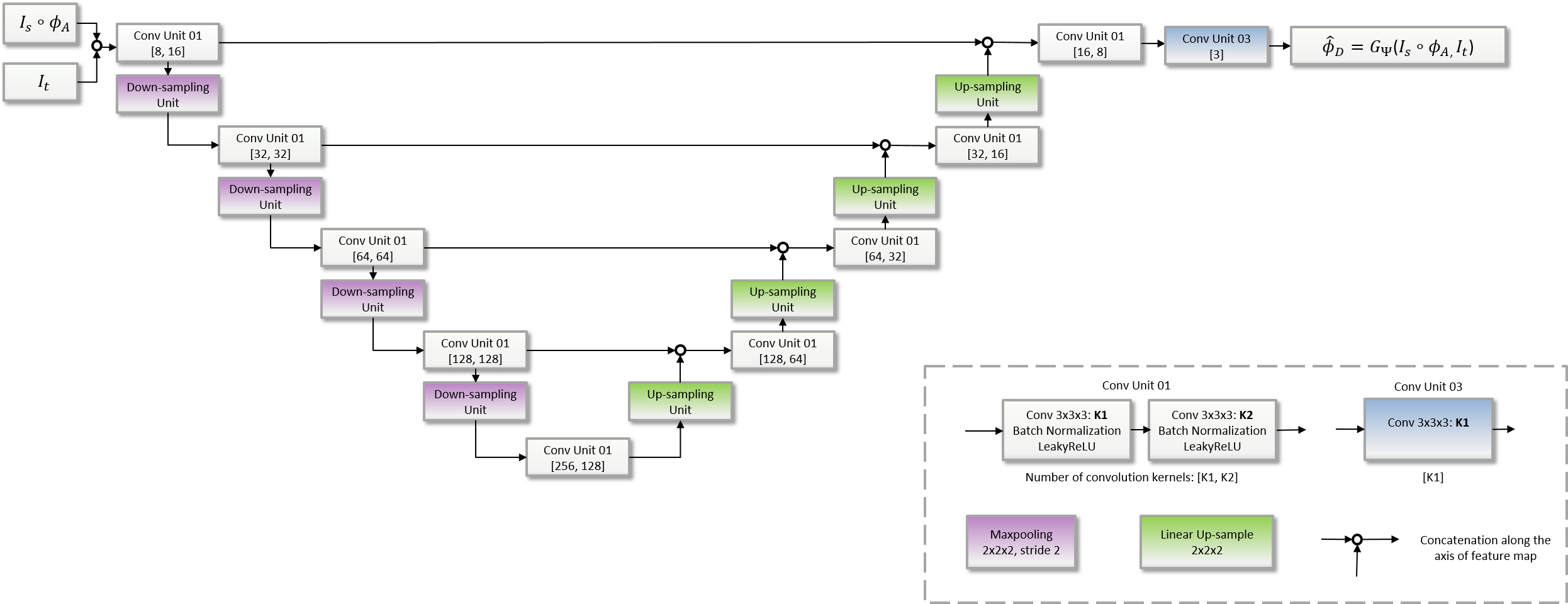}
\caption{The network architectures of $\mathcal{G}_{\Psi}$ module used in the present study. The target image ($I_t$) and affine-aligned source image ($I_s \circ \boldsymbol{\phi}_A$) are used as the input to predict non-rigid deformation ($\boldsymbol{\phi}_D$), which can subsequently lead to a composite displacement field ($\boldsymbol{\phi}$) as shown in Figure \ref{fig1}. The number of convolution kernels used in Conv Unit 01 is denoted as [K1, K2], and that for Conv Unit 03 as [K1].}\label{fig:s2}
\end{figure*}

The network architectures of $\mathcal{F}_{\Theta}$ and $\mathcal{G}_{\Psi}$ are illustrated in Figure \ref{fig:s1} and Figure \ref{fig:s2}, respectively. The image size of the input and output for $\mathcal{F}_{\Theta}$ used in the present study were ($112\times208\times112\times6$) and ($112\times208\times112\times3$) voxels. That of $\mathcal{G}_{\Psi}$ were ($112\times208\times112\times2$) and ($112\times208\times112\times3$) voxels.

The procedure of simultaneous optimization is summarized with the pseudo code in Algorithm \ref{algorithm1}. 
\begin{algorithm}[!h]
\caption{Simultaneous optimization}\label{algorithm1}
\SetAlgoLined
\SetKwInput{KwInit}{Initialization}
\SetKwInput{KwPara}{Parameters}
\KwIn{$\{ \mathcal{S}_s^i,\, \mathcal{S}_t^i,\, I_s^i\,, I_t^i, \boldsymbol{\phi}_A^i \}_{i=1}^N$}
\KwPara{$\boldsymbol{\Theta}, \boldsymbol{\Psi}$}
\KwOut{$\{ \hat{\mathcal{S}}_s^i,\, \hat{\mathcal{S}}_s^i \circ \hat{\boldsymbol{\phi}}^i,\, I_s^i \circ \hat{\boldsymbol{\phi}^i},\, \hat{\boldsymbol{\phi}^i},\, \hat{\boldsymbol{\phi}^i}_D\}_{i=1}^N $}
\KwInit{}
$\text{lr}, \text{decay\_factor} $\tcp*[f]{Initial and decay ratio of learning rate}\\
$\boldsymbol{\Theta}, \boldsymbol{\Psi} \sim \text{GlorotUniform}$ \tcp*[f]{Kernel initialization} \\

\For(){$\text{\upshape number of training iterations}$}{

\text{shuffle}(\textbf{Input} ) \\
    \For(){$i=0$ \KwTo $N$}{
    
    $\hat{\mathcal{S}}_s^i = \mathcal{F}_{\Theta}\big(I_s^i\big)$ \tcp*[f]{Segmentation, Eq.\ref{eq1}}\\
    $\hat{\boldsymbol{\phi}}_D^i = \mathcal{G}_{\Psi} \big(I_t^i,\, I_s^i \circ \boldsymbol{\phi}_A^i\big)$ \tcp*[f]{Registration, Eq.\ref{eq4}}\\
    $\hat{\boldsymbol{\phi}^i} = \boldsymbol{\phi}_A^i \circ \hat{\boldsymbol{\phi}}_D^i$ \tcp*[f]{Transform composition}\\
    $ \hat{I}_t^i = I_s^i \circ \hat{\boldsymbol{\phi}^i} $ \tcp*[f]{Image warp}\\
    
    \tcc{Dependency on both tasks}
    $ \hat{\mathcal{S}}_t^i = \hat{\mathcal{S}}_s^i \circ \hat{\boldsymbol{\phi}^i}$ \tcp*[f]{Segmentation warp}
    
    $\mathcal{L}^i = \mathcal{L}_{seg} (\mathcal{S}_s^i,\, \hat{\mathcal{S}}_s^i) + \alpha \mathcal{L}_{reg} (I_t^i,\, \hat{I}_t^i) + \beta \mathcal{L}_{def} (\hat{\boldsymbol{\phi}}_D^i)$\\
    $\hfill + \gamma \mathcal{L}_{com} (\mathcal{S}_t^i,\, \hat{\mathcal{S}}_t^i) $
    \tcp*[f]{Segis-Net objective, Eq.\ref{eq9}}\\

    \tcc{Simultaneous optimization of parameters}
    $\boldsymbol{\Theta}, \boldsymbol{\Psi} \leftarrow $ \textbf{Adam} ($\mathcal{L}^i$, lr,  $\boldsymbol{\Theta}, \boldsymbol{\Psi}$)\\
}
\tcc{Custom condition of learning rate decay}
\If{$\text{\upshape learning rate decay}$}{
$lr \leftarrow lr \times$ \text{decay\_factor}}

\KwRet{$\boldsymbol{\Theta}, \boldsymbol{\Psi}$} \tcp*[f]{Return parameters per epoch}
}
\end{algorithm}

\begin{table}[h!]
\centering
\small
 \begin{tabular}{ c|c|c|c|c } 
  \multicolumn{2}{c|}{} & Classical & CNN & Segis-Net \\
  \cline{1-5} 
  \multirow{6}{*}{\rotatebox[origin=c]{90}{Figure~\ref{fig2}}} 
  & CGC\_L & - & 0.76 $\pm$ 0.06 & 0.76 $\pm$ 0.06 \\ 
  & CGC\_R & - & 0.76 $\pm$ 0.06 & 0.76 $\pm$ 0.06 \\
  & CGH\_L & - & 0.76 $\pm$ 0.07 & 0.76 $\pm$ 0.07 \\
  & CGH\_R & - & 0.77 $\pm$ 0.09 & 0.76 $\pm$ 0.09 \\
  & FMA    & - & 0.76 $\pm$ 0.05 & 0.76 $\pm$ 0.05 \\
  & FMI    & - & $0.68 \pm 0.09$ & 0.67 $\pm$ 0.09 \\
  \cline{1-5}
  \multirow{6}{*}{\rotatebox[origin=c]{90}{Figure~\ref{fig3}}} 
  & CGC\_L & 0.73 $\pm$ 0.10 & 0.71 $\pm$ 0.07 & 0.73 $\pm$ 0.08  \\ 
  & CGC\_R & 0.73 $\pm$ 0.10 & 0.69 $\pm$ 0.06 & 0.73 $\pm$ 0.07 \\
  & CGH\_L & 0.75 $\pm$ 0.11 & 0.75 $\pm$ 0.10 & 0.77 $\pm$ 0.09 \\
  & CGH\_R & 0.75 $\pm$ 0.11 & 0.75 $\pm$ 0.10 & 0.76 $\pm$ 0.10\\
  & FMA    & 0.72 $\pm$ 0.10 & 0.71 $\pm$ 0.08 & 0.74 $\pm$ 0.09 \\
  & FMI    & 0.74 $\pm$ 0.08 & 0.75 $\pm$ 0.08 & 0.76 $\pm$ 0.08 \\
  \cline{1-5}    
  \multirow{6}{*}{\rotatebox[origin=c]{90}{Figure~\ref{fig4}}} 
  & CGC\_L & 0.68 $\pm$ 0.06 & 0.82 $\pm$ 0.04 & 0.83 $\pm$ 0.04\\ 
  & CGC\_R & 0.68 $\pm$ 0.06 & 0.82 $\pm$ 0.06 & 0.83 $\pm$ 0.04 \\
  & CGH\_L & 0.66 $\pm$ 0.08 & 0.81 $\pm$ 0.05 & 0.82 $\pm$ 0.05 \\
  & CGH\_R & 0.66 $\pm$ 0.09 & 0.81 $\pm$ 0.05 & 0.81 $\pm$ 0.05 \\
  & FMA    & 0.69 $\pm$ 0.06 & 0.84 $\pm$ 0.03 & 0.87 $\pm$ 0.02 \\
  & FMI    & 0.57 $\pm$ 0.09 & 0.77 $\pm$ 0.07 & 0.81 $\pm$ 0.05  \\
  \cline{1-5}  
  \multirow{6}{*}{\rotatebox[origin=c]{90}{Figure~\ref{fig5}(a)}}
  & CGC\_L & 0.68 $\pm$ 0.07 & 0.82 $\pm$ 0.04 & 0.83 $\pm$ 0.03 \\
  & CGC\_R & 0.68 $\pm$ 0.07 & 0.81 $\pm$ 0.04 & 0.82 $\pm$ 0.04 \\
  & CGH\_L & 0.65 $\pm$ 0.10 & 0.77 $\pm$ 0.07 & 0.79 $\pm$ 0.06 \\
  & CGH\_R & 0.66 $\pm$ 0.09 & 0.78 $\pm$ 0.05 & 0.79 $\pm$ 0.05 \\
  & FMA    & 0.72 $\pm$ 0.06 & 0.85 $\pm$ 0.03 & 0.87 $\pm$ 0.03 \\
  & FMI    & 0.64 $\pm$ 0.08 & 0.79 $\pm$ 0.08 & 0.82 $\pm$ 0.06 \\
  \cline{1-5} 
  \multirow{6}{*}{\rotatebox[origin=c]{90}{Figure~\ref{fig5}(b)}}
  & CGC\_L & 11 $\pm$ 8.8\%  & 5.1 $\pm$ 3.6\% & 4.8 $\pm$ 4.1\%  \\
  & CGC\_R & 11 $\pm$ 8.9\%  & 5.4 $\pm$ 4.5\%  & 4.5 $\pm$ 3.9\% \\
  & CGH\_L & 11 $\pm$ 9.8\%  & 7.9 $\pm$ 7.2\%  & 7.3 $\pm$ 5.6\%  \\
  & CGH\_R & 9.8 $\pm$ 7.3\% & 7.6 $\pm$ 7.5\%  & 7.0 $\pm$ 5.8\% \\
  & FMA    & 6.6 $\pm$ 5.9\% & 4.9 $\pm$ 3.6\%   & 3.4 $\pm$ 2.6\%  \\
  & FMI    & 7.9 $\pm$ 6.5\% & 8.5 $\pm$ 8.6\% & 6.0 $\pm$ 6.2\% \\
  \cline{1-5}   
  \multirow{6}{*}{\rotatebox[origin=c]{90}{Figure~\ref{fig5}(c)}} 
  & CGC\_L & 4.7 $\pm$ 4.0\% & 3.0 $\pm$ 2.2\% & 2.8 $\pm$ 2.2\% \\
  & CGC\_R & 4.6 $\pm$ 3.4\% & 3.3 $\pm$ 2.6\% & 3.4 $\pm$ 2.4\% \\
  & CGH\_L & 7.3 $\pm$ 5.5\% & 4.7 $\pm$ 3.9\% & 5.2 $\pm$ 4.0\% \\
  & CGH\_R & 6.3 $\pm$ 4.7\% & 4.3 $\pm$ 3.8\% & 4.6 $\pm$ 4.1\% \\
  & FMA    & 1.9 $\pm$ 1.5\% & 1.9 $\pm$ 1.4\% & 2.0 $\pm$ 1.6\% \\
  & FMI    & 2.7 $\pm$ 1.9\% & 2.7 $\pm$ 2.1\% & 2.3 $\pm$ 1.7\% \\
  \cline{1-5}   
  \multirow{6}{*}{\rotatebox[origin=c]{90}{Figure~\ref{fig5}(d)}}
  & CGC\_L & 1.9 $\pm$ 1.8\% & 1.7 $\pm$ 1.5\% & 1.6 $\pm$ 1.4\% \\
  & CGC\_R & 1.5 $\pm$ 1.2\% & 1.5 $\pm$ 1.2\% & 1.4 $\pm$ 1.1\% \\
  & CGH\_L & 3.1 $\pm$ 4.9\% & 2.1 $\pm$ 1.6\% & 2.2 $\pm$ 1.6\% \\
  & CGH\_R & 2.3 $\pm$ 1.9\% & 1.7 $\pm$ 1.5\% & 1.6 $\pm$ 1.4\% \\
  & FMA    & 1.4 $\pm$ 1.1\% & 1.1 $\pm$ 1.1\% & 1.1 $\pm$ 0.9\% \\
  & FMI    & 1.1 $\pm$ 0.8\% & 1.1 $\pm$ 0.9\% & 1.0 $\pm$ 0.8\% \\
  \cline{1-5} 
 \end{tabular}
\caption{Results overview for Figure 3-5. Figure~\ref{fig2}, Segmentation accuracy; Figure~\ref{fig3}, Spatial correlation (SC) similarity; Figure~\ref{fig4}, Spatio-temporal consistency of segmentation (STCS); Figure~\ref{fig5} (a), Segmentation reproducibility; Figure~\ref{fig5} (b), Volume Reproducibility; Figure~\ref{fig5} (c), FA reproducibility; Figure~\ref{fig5} (d), MD reproducibility.}\label{tab3}
\end{table}

\end{document}